\pdfoutput=1
\documentclass{article}

\usepackage{microtype}
\usepackage{bm}
\usepackage{listings}
\usepackage{fancyvrb}
\usepackage{multirow}
\usepackage{xcolor}
\usepackage{makecell}
\usepackage{bbm}
\usepackage{graphicx}
\usepackage{subfigure}
\usepackage{url}
\usepackage{booktabs} 

\usepackage[colorlinks, linkcolor=blue, citecolor=green, urlcolor=red]{hyperref}

\usepackage[nolinenum]{icml2025}

\usepackage{amsmath}
\usepackage{amssymb}
\usepackage{mathtools}
\usepackage{amsthm}

\usepackage[capitalize,noabbrev]{cleveref}

\theoremstyle{plain}

\theoremstyle{definition}

\theoremstyle{remark}

\usepackage[textsize=tiny]{todonotes}

\icmltitlerunning{Fitting Unknown Number of Hyperplanes with Manifold Optimization}

\begin{document}

\twocolumn[
\icmltitle{Fitting Unknown Number of Hyperplanes with Manifold Optimization}

\begin{icmlauthorlist}
\icmlauthor{Zhiqin Cheng}{polyU}
\icmlauthor{Yu Zhan}{sustech}
\icmlauthor{Mingjin Zhang}{polyU}
\icmlauthor{Lingbo Liu}{pcl}
\icmlauthor{Liang Lin}{sysu}

\end{icmlauthorlist}
\icmlaffiliation{sustech}{Department of Electronic and Electrical Engineering, Southern University of Science and Technology, ShenZhen, China}
\icmlaffiliation{polyU}{Department of Computing, The Hong Kong Polytechnic University, Kowloon, Hong Kong}
\icmlaffiliation{sysu}{School of Comuputer Science and Engineering, Sun Yat-Sen University, GuangZhou, China}
\icmlaffiliation{pcl}{Research Institute of Multiple Agents and Embodied Intelligence, Pengcheng Laboratory, ShenZhen, China}

\icmlcorrespondingauthor{Zhiqin Cheng}{eecs.zqcheng@outlook.com}

\icmlkeywords{Machine Learning, ICML}

\vskip 0.3in
]

\printAffiliationsAndNotice{}  

\begin{abstract}

Fitting an unknown number of hyperplanes to data is a fundamental yet challenging problem in machine learning, characterized by its non-convexity, non-differentiability, and unknown model order. Existing approaches often struggle with local optima or lack geometric consistency. To address these limitations, we propose a novel framework based on Manifold Optimization. We reformulate the problem as an unsupervised learning task on the unit sphere manifold $\mathcal{S}^{\textbf{dim}-1}$. This formulation effectively handles the non-convex constraints and linearizes the distance measurement, rendering the gradient descent tractable. We propose a Two-Stage Manifold Optimization algorithm. In Phase I, we employ a Riemannian Expectation-Maximization process with a heavy-tailed kernel to robustly estimate posterior probabilities, effectively resolving the ambiguities of point distribution between intersecting hyperplanes. In Phase II, upon convergence of the soft estimates, the probabilistic weights degenerate into hard matching, generating a precise local optimum that strictly satisfies the geometric definition. Furthermore, we introduce a projected density estimation strategy for initialization to facilitate global convergence by significantly reducing the feature description space and search complexity. Extensive experiments demonstrate that our method outperforms state-of-the-art baselines in both geometric accuracy and robustness. Our code is available at \href{https://github.com/aaronworry/HyperplanesFitting}{https://github.com/aaronworry/HyperplanesFitting}.
\end{abstract}

\section{Introduction}
\label{introduction}

In recent years, many problems are nonconvex, which remain challenges to design algorithms to solve them\cite{Zhichao2025first}, as the gradient descent method usually gets a local optimum. To overcome it, researchers mainly proposed two methodologies: 1) escaping from local optimum, such as simulated annealing\cite{Kirkpatrick1983OptimizationBS}, gradient descent with momentum\cite{GDM1986hinton}, and adaptive moment estimation\cite{kingma2017adammethodstochasticoptimization}. 2) Selecting multiple initial values and perform parallel computing, then pick the optimal. This category includes multi-start optimization\cite{boender1982stochastic}, genetic algorithms\cite{mitchell1996introduction}, and particle swarm optimization\cite{kennedy1995particle}. However, the former may not get the global optimum while the latter is computationally expensive, especially when the state space of the solution is large. In this work, we study fitting unknown number of hyperplanes, which comes from multisource hyperplanes location\cite{vctor2021On, vctor2018location}, a mixed-integer programming problem. To solve the problem, we propose a methodology based on manifold optimization (introduced in Appendix.\ref{pre_manifold}), which is used in many learning problems where a geometry that better matches the structure of the data\cite{10840334, huang2025an, chami2019hyperbolicgraphconvolutionalneural, shukla2019geometrydeepgenerativemodel}. Moreover, we proposed a method to get an available initial value. 

\begin{figure}[t]
\vskip 0.2in
\begin{center}
\includegraphics[width=0.45\textwidth]{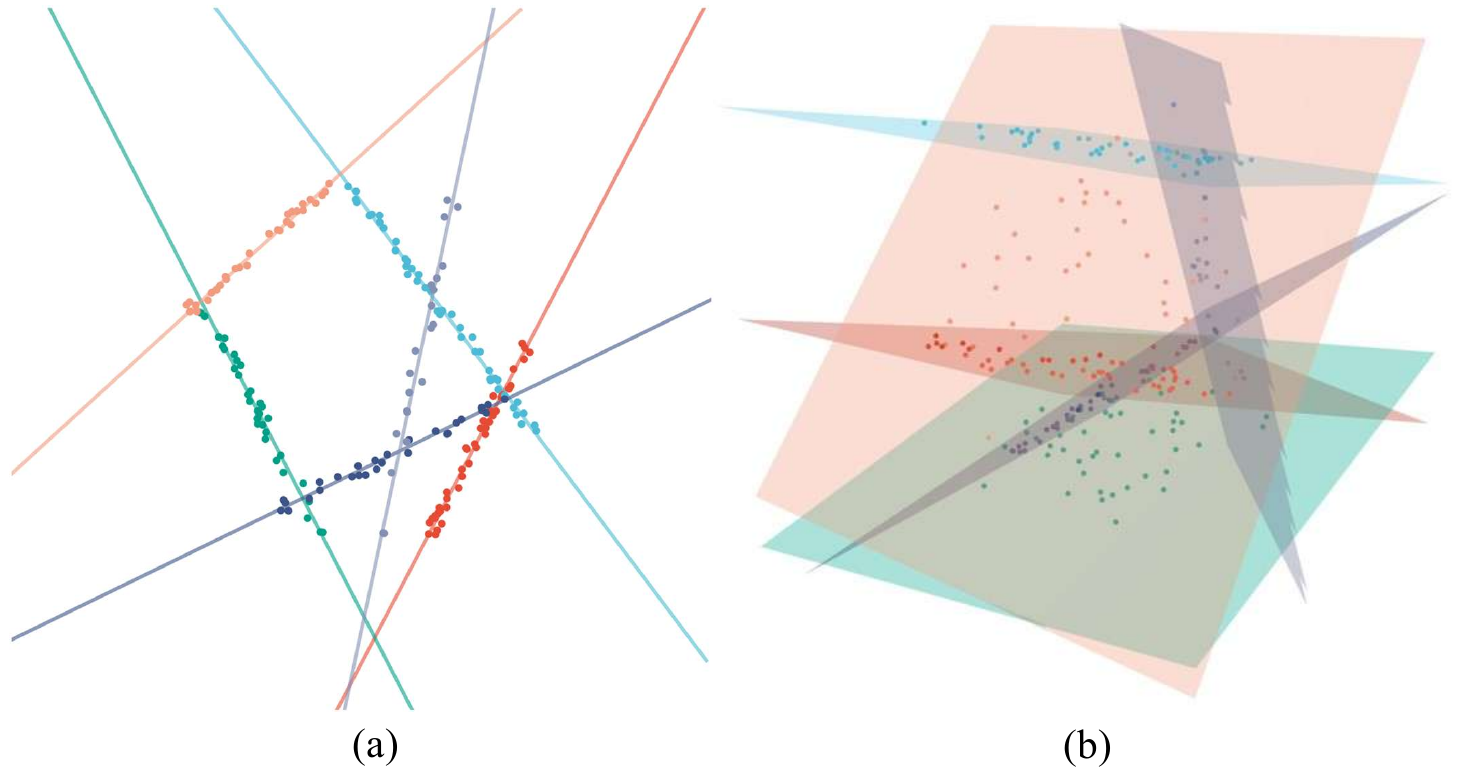}
\vspace{-7mm}
\caption{Examples of hyperplane fitting on 2D and 3D data containing six hyperplanes.}
\label{fig:hyperplane_fitting_instance}
\end{center}
\vspace{-5mm}
\end{figure}

Fitting unknown number of hyperplanes is a non-convex and non-differentiable problem. Once the number of hyperplanes is determined, it can be regarded as an unsupervised learning problem, i.e., fuzzy clustering\cite{10542191, Ramos-Carreo2024, Xue2025Tensorized}. However, those algorithms failed to fit hyperplanes because there are no clear boundaries between clusters, making it difficult to determine which cluster a point belongs to. And the parameters of hyperplanes are difficult to estimate, as the distance computation between points and hyperplanes is nonlinear, making the estimation of parameters non-convex. 


Deep learning methods are difficult to handle such problems, where both the input and output dimensions vary dynamically. Although some researchers have attempted to solve such problems in an engineering manner — for example, describing input data in the form of images\cite{leiber2024dying, miklautz2025breaking, yang2017towards, Miklautz2020deep} or limiting the dimensions of inputs and the quantity of outputs\cite{qi2017pointnet, li2018pointcnn, wu2024point} — these methods have significant limitations. The former can only handle 2D scenarios, and converting data into images leads to information loss. The latter requires redesigning neural networks for different input and output dimensions and training distinct neural networks on separate datasets consumes additional computational resources. Moreover, neural networks face challenges when a non-differentiable objective function and non-convex constraints are introduced\cite{soudry2024implicitbiasgradientdescent, raissi2019physics}, as the computed gradient may not guide convergence, and the parameters updated via gradient descent may fail to satisfy the conditions of non-convex constraints, respectively.

To address these challenges, we present a Two-Stage Manifold Optimization algorithm. In the first stage, we leverage a Riemannian Expectation-Maximization framework with a heavy-tailed kernel for robust posterior probability estimation, resolving point distribution ambiguities across intersecting hyperplanes. In the second stage, we convert convergent soft estimates into hard probabilistic matching, yielding an accurate local optimum that strictly satisfies geometric definitions. We also propose a projected density estimation initialization strategy, which reduces feature space and search complexity to approach global optima. The contributions of this work are:
\begin{itemize}
 \vspace{-1.em}
\setlength{\itemsep}{0.2em}    
\setlength{\parskip}{0.0em}  
\item We reformulate the whole problem into a clustering problem to eliminate the non-differentiability. 
\item We apply the unit normal vector to describe the hyperplanes, reducing the state space of the hyperplane(Sec.\ref{sampling}) and linearizing the distance calculation between points and hyperplanes.
\item We use manifold $\mathcal{S}$ to represent the feature of hyperplanes and propose algorithms based on manifold optimization to solve the problem, which can eliminate the non-convex constraints introduced by reformulated problem and manifold $\mathcal{S}$.
\item We proposed a method to determine potential hyperplanes in points, which can help our method to get global optimum.
 \vspace{1.em}
\end{itemize}



\section{Preliminaries and Problem} \label{problem}

\subsection{Hyperplane}
A hyperplane $h$ in $\textbf{dim}$-dimensional space is a sub-space $\mathbb{R}^{\textbf{dim}-1} \subset \mathbb{R}^{\textbf{dim}}$, which can be represented by a coefficient $\bm{a} \in \mathbb{R}^{\textbf{dim}}$ and offset $b \in \mathbb{R}$:
\begin{equation}
h(\bm{a}, b) = \{x \in \mathbb{R}^{\textbf{dim}} | \textbf{a}^Tx + b = 0\}
\end{equation} For example, a hyperplane in 2D space is a line, while it is a plane in 3D space, respectively.

Fitting a set of points with one hyperplane can be regarded as a problem of finding the center of a cluster that consists of points. And the center of this cluster $C$ is hyperplane $h(\bm{a}, b)$ satisfies:
\begin{equation}
\bm{a},b = \arg\min_{\bm{a},b} \sum_{x_i \in C} \frac{\lvert \bm{a}^T x_i + b \rvert}{\lVert \bm{a} \rVert_2}
\end{equation}

\subsection{Problem Definition} \label{problem_definition}
This work focuses on fitting a set of points with an unknown number of hyperplanes, which can be formulated as below.

Given a set of points $\mathcal{X} = \{x_1, \cdots, x_n\}$ with $\forall i \le n,x_i \in \mathbb{R}^\textbf{dim}$, our goal is finding $m$ hyperplanes to fit these $n$ points, i.e. minimizing the sum of the distances from these points to the nearest hyperplane. The objective function can be summarized as 
\begin{equation} \label{eq:original_obj}
\min \quad \sum_{i=1}^n \min_{\forall j \le m}\frac{\lvert \bm{a}_j^{T}x_i+b_j \rvert}{{\lVert \bm{a}_j \rVert}_2}
\end{equation}
where $\bm{a}_j, b_j$ are coefficient and offset of the $j$-th hyperplane. And those hyperplanes form a set $\mathcal{H} = \{h_1, \cdots, h_m\}$ while $h_j=(\bm{a}_j, b_j)$. It is noted that in \textbf{dim}-dimensional space, $\bm{a}_j \in \mathbb{R}^\textbf{dim}, b_j \in \mathbb{R}$.

\section{Method} \label{method}
As the problem defined before is non-convex and non-differentiable, existing solvers can not perform well (Appendix.\ref{compare_solver}). Therefore, to solve the problem, we first reformulate the problem, then introduce the method to solve it.

\subsection{Optimization Problem Formulation}
The problem mentioned in Sec.\ref{problem_definition} can be regarded as a clustering problem: Finding $m$ clusters $\{C_j|j=1,2,\cdots,m\}$ where $C_k \cap_{k \neq l} C_l = \varnothing$ and $\cup_{j=1}^m C_j = \mathcal{X}$. Moreover, the center of $j$-th cluster is hyperplane $h_j = (\bm{a}_j, b_j)$ and the distance metric
\begin{equation}
D(x_i, C_j) = \frac{\lvert \bm{a}_j^{T}x_i+b_j \rvert}{{\lVert \bm{a}_j \rVert}_2}
\end{equation} Thus, the problem defined in Sec.\ref{problem_definition} can be formulated as 
\begin{equation} \label{eq:optimi}
\begin{gathered}
\min \sum_{j=1}^{m} \sum_{x_i \in C_j}D(x_i, C_j) \\
\begin{matrix}
\text { s. t. } & \bigcup_{j=1}^{m}C_j = \{x_1, x_2, \cdots, x_n\} \\
 & C_l \cap C_k = \varnothing & \forall l \ne k
\end{matrix}
\end{gathered}
\end{equation} 

To simplify the computation of the distance metric, we introduce a normal vector $n_j$ and a positive distance parameter $d_j \ge 0$ to represent the hyperplane (center of the cluster). Therefore, the hyperplane and distance metric can be represented as $h_j = (\textbf{n}_j, d_j)$ and 
\begin{equation}
D(x_i, C_j) = \lvert \textbf{n}_j \cdot x_i - d_j \rvert
\end{equation} respectively, as well as new constraints are introduced:
\begin{equation} \label{eq:non-linear_constr}
\lVert \textbf{n}_j \rVert_2 = 1, \quad \forall 1\le j \le m
\end{equation}

Eq.(\ref{eq:non-linear_constr}) is intractable for classic numerical methods and optimization solvers, which is shown in Appendix~\ref {compare_solver}, due to its non-convexity. Therefore, it is necessary to propose a new methodology to handle this kind of problem.

Combined with all equations mentioned before, we reformulate the optimization problem clearly.
\begin{equation} \label{eq:optimization_problem}
\begin{gathered}
\min_{h_1, \cdots, h_m} \sum_{j=1}^{m} \sum_{x_i \in C_j}\lvert \textbf{n}_j \cdot x_i - d_j \rvert \\
\begin{matrix}
\text { s. t. } & \bigcup_{j=1}^{m}C_j = \{x_1, x_2, \cdots, x_n\} \\
 & C_l \cap C_k = \varnothing & \forall l \ne k \\
 & \lVert \textbf{n}_j \rVert_2 = 1& \forall 1 \le j \le m
\end{matrix}
\end{gathered}
\end{equation} where center of cluster $C_j$ is hyperplane $h_j = (\textbf{n}_j, d_j)$. 

\subsection{Manifold Representation}

To circumvent the non-convex equality constraints defined in eq.\ref{eq:optimization_problem} during optimization, we introduce the Riemannian manifold structure.

\textbf{Manifold}: The constraint $\lVert \textbf{n} \rVert_2 = 1$ can be released by defining a Riemannian manifold 
\begin{equation}
\label{eq:manifold}
\mathcal{S} = \{x|~\lVert x \rVert_2 = 1\}
\end{equation}
to represent the normal vector of hyperplanes.

\textbf{Tangent Space}: When updating variables on a manifold, the gradient of the objective function should be projected to the manifold's tangent space defined on the last value. For the manifold $\mathcal{S}$, the tangent space defined on point $x$ is described by
\begin{equation}
\label{eq:tangent_space}
\mathcal{T}_x\mathcal{S} = \{v |~x \cdot v = 0\}.
\end{equation}
Consequently, the gradient of the objective function projected to the tangent space could be linearly represented by a set of linearly independent vectors $v \in \mathcal{T}_x\mathcal{S}$. 

\textbf{Retraction}: To guarantee that the updated value is on the manifold $\mathcal{S}$, the retraction function is defined as 
\begin{equation}
\label{eq:retraction}
\mathcal{R}_x(v) = \frac{x+v}{\lVert x+v \rVert_2}
\end{equation}
where $v \in \mathcal{T}_x\mathcal{S}$. The value of $v$ represents the step size and gradient when updating the variable at $x$. Therefore, the updated $\hat{x}$ can be computed by $\hat{x} = \mathcal{R}_{x}(v)$.

\subsection{Manifold Optimization} \label{manifold_optimization}
In our problem, solving the normal vector $\textbf{n}_j$ of the objective function can be relaxed to minimizing the variance of the distances from the points to the center(hyperplane) of the cluster, while the mean value is the parameter $d_j$. The optimal normal vectors of these two objective functions are very close, because the distances from the points to the ground-truth hyperplane are very small. In this part, we mainly introduce the optimization part of our methods, while the clustering and initial values of its centers are demonstrated later.

Once the initial values for the $m$ hyperplanes are determined, the problem (Eq.\ref{eq:optimization_problem}) is solved using Algorithm \ref{alg:manifold_optimization}. We employ a mixture-model-based approach to compute an assignment weight matrix (lines 4-15). Points are then assigned to specific clusters based on the maximum weight (line 17). Finally, the hyperplane parameters for each cluster are refined using the points assigned to it (lines 18-21).

\begin{algorithm}[t]
   \caption{Two-Stage Manifold Optimization Framework}   \label{alg:manifold_optimization}
    \begin{algorithmic}[1]
   \STATE {\bfseries Input:} a set of points $\mathcal{X} = \{x_1, \cdots, x_n\}$, a set of initial hyperplanes $\hat{\mathcal{H}} = \{h_1, \cdots, h_m\}$.
   \STATE {\bfseries Output:}  a set of updated hyperplanes $\mathcal{H} = \{h_1, \cdots, h_m\}$.
   \STATE  \textcolor{blue}{// \textit{Phase I: Soft Optimization (Riemannian EM)}}
   \STATE Weight matrix $w_{ij} = 0$ 
    \REPEAT
    \STATE $w_{ij}^{old} = w_{ij}$
    
    \STATE \textcolor{blue}{// E-step: Update posterior probabilities using inverse-square kernel}
    \STATE Computing weight matrix $w_{ij} = \frac{\frac{1}{\lvert \textbf{n}_j \cdot x_i - d_i \rvert^2}}{\sum_{k=1}^{m}\frac{1}{\lvert \textbf{n}_k \cdot x_i - d_k \rvert^2}}$ 
    \STATE \textcolor{blue}{// M-step}
    \FOR{$j=1$ {\bfseries to} $m$}
    \STATE $\textbf{n}_j \gets \texttt{ManifoldOptimization}(\mathcal{X}, w_j) $
    \STATE Computing $d_j=\frac{\sum_{i=1}^{n}w_{ij}\textbf{n}_j\cdot x_i}{\sum_{i=1}^n w_{ij}}$
    \ENDFOR

    \STATE Computing $\Delta w_{ij} = w_{ij} - w_{ij}^{old}$
    \UNTIL{$\max{\lvert \Delta w_{ij} \rvert} <  tolerance$}
    \STATE \textcolor{blue}{\textit{// Phase II: Hard Refinement}}
    \STATE Determining the cluster $C_1, \cdots, C_m$ based on $w_{ij}$, i.e. point $x_i$ is in $C_j$ while $w_{ij}$ is the maximum in $w_i$.

    \FOR{$j=1$ {\bfseries to} $m$}
    \STATE $\textbf{n}_j \gets \texttt{ManifoldOptimization}(C_j, \mathbbm{1}) $
    \STATE Computing $d_j=\frac{\sum_{x_i \in C_j} \textbf{n}_j\cdot x_i}{\textbf{card}(C_j)}$, \textbf{card} means the number of elements in set.
    \ENDFOR
    
    \STATE {\bfseries return} $\mathcal{H}$
\end{algorithmic}
\end{algorithm}

To enhance the distinction between hyperplanes and eliminate ambiguity of point distribution, we adopt a probabilistic perspective. Instead of standard Gaussian assumptions, we model the probability that a point belongs to a hyperplane using a heavy-tailed kernel (specifically, the inverse-square distance). This formulation allows us to derive the posterior probabilities (weights) for the Expectation-Maximization (EM) process.

Since the non-differentiability of the absolute value calculation of eq.\ref{eq:optimization_problem} near zero, it is difficult to get a suitable gradient. To overcome it, we replace it with a quadratic function and use optimization to minimize the variance of the distance. Therefore, the function \texttt{ManifoldOptimization} is used to update the normal vector:
\begin{equation} \label{eq:manifold_optimization}
\begin{gathered}
\begin{matrix}
& \min_{\textbf{n}} \frac{\sum_i w_i(\textbf{n}\cdot x_i-\frac{\sum_i w_i\textbf{n}\cdot x_i}{\sum_i w_i})^2}{\sum_i w_i} \\
\text { s. t. } 
 & \lVert \textbf{n} \rVert_2 = 1
\end{matrix}
\end{gathered}
\end{equation} where $x_i$ and $w_i$ are points and weights respectively. It is noted that this formula is a trade-off to obtain a more stable solution. Meanwhile, it also reduces the consideration of distance $d$. And the problem eq.\ref{eq:manifold_optimization} can be solved through algo.\ref{alg:solution_for_manifold}. To compute the Riemannian gradient on $\textbf{n}$, we need to compute the Euclidean gradient $\textbf{e} = \left.\frac{\mathrm{d}f}{\mathrm{d}\textbf{n}}\right|_{\textbf{n}}$, then projecting it to tangent space defined on point $\textbf{n}$. Here we introduce the projection, while the euclidean gradient is introduced latter as there are two function in algo.\ref{alg:manifold_optimization}, i.e. line 11 in Sec.\ref{method_A} and line 19 in Sec.\ref{method_B} respectively.

\textbf{Projection}: This operation can project the vector to the tangent space. Since the norm of $\textbf{n}$ is 1, the projection vector in tangent space defined on $\textbf{n}$ can be calculated by:
\begin{equation}
\textup{Proj}_{\textbf{n}}(\textbf{e}) = \textbf{e} - (\textbf{e}\cdot \textbf{n})\textbf{n}
\end{equation}

\begin{algorithm}[t]
   \caption{ ~\texttt{ManifoldOptimization}}   \label{alg:solution_for_manifold}
    \begin{algorithmic}[1]
   \STATE Chose an initial value $\textbf{n}_0 \in \mathcal{S}$

   \STATE $\textbf{n} = \textbf{n}_0$ 

   \REPEAT

   \STATE Computing the Riemannian gradient of the objective function: $\eta = \textup{grad}~ f(\textbf{n})$

    \STATE Computing the step size $\alpha$ through line search.

    \STATE Updating the normal vector through retraction: $\textbf{n} = \mathcal{R}_\textbf{n}(-\alpha \eta) $
   
    \UNTIL{convergence}

    \STATE {\bfseries return} $\textbf{n}$
\end{algorithmic}
\end{algorithm}

\subsubsection{Phase I: Soft Optimization (Riemannian EM)} \label{method_A}
This phase corresponds to lines 4-15 in Algo.\ref{alg:manifold_optimization}. Specifically, we implement the E-step by defining the likelihood using an inverse-square metric ($1/D^2$). Compared to exponential decay (Gaussian), this heavy-tailed kernel imposes stronger penalties on close inliers while maintaining long-range attraction for unassigned points, providing robustness in the early stages of optimization. By treating the assignment of points to hyperplanes as a soft probability (weight $w_{ij}$), this phase effectively explores the solution space and avoids premature convergence to local optima. It can be viewed as the Expectation-Maximization (EM) process on the manifold.

For line 11 in algo.\ref{alg:manifold_optimization}, the euclidean gradient of objective function in eq.\ref{eq:manifold_optimization} can be computed through:
\begin{equation}
\left.\frac{\mathrm{d}f}{\mathrm{d}\textbf{n}}\right|_{\textbf{n}} = \frac{\sum_{i=1}^n 2w_i(\textbf{n}\cdot x_i - \frac{\sum_{i=1}^n w_i\textbf{n}\cdot x_i}{\sum_{i=1}^n w_i})(x_i - \frac{\sum_{i=1}^n w_i x_i}{\sum_{i=1}^n w_i})}{\sum_{i=1}^{n}w_i}
\end{equation}

\subsubsection{Phase II: Hard Refinement} \label{method_B}
This phase corresponds to lines 17-21 in Algo.\ref{alg:manifold_optimization}. After the probabilistic assignments stabilize, we perform a deterministic "hard" assignment to fix the clusters. Then, an unweighted manifold optimization is performed within each cluster to fine-tune the geometric parameters $(\textbf{n}_j, d_j)$ to maximize precision.

For line 19 in algo.\ref{alg:manifold_optimization}, the euclidean gradient of objective function in eq.\ref{eq:manifold_optimization} can be computed through:
\begin{equation}
\left.\frac{\mathrm{d}f}{\mathrm{d}\textbf{n}}\right|_{\textbf{n}} = \frac{\sum_{x_i \in C} 2(\textbf{n}\cdot x_i - \frac{\sum_{x_i \in C} \textbf{n}\cdot x_i}{\textbf{card}(C)})(x_i - \frac{\sum_{x_i \in C} x_i}{\textbf{card}(C)})}{\textbf{card}(C)}
\end{equation}

\vspace{0.5cm}

In summary, Algo.\ref{alg:manifold_optimization} presents a coherent framework comprising two phases: Phase I performs soft optimization via Riemannian EM to address ambiguity, followed by Phase II, which performs hard refinement to achieve geometric precision.

\section{Sampling on Manifold} \label{sampling}

To employ the optimization framework described in Sec.\ref{manifold_optimization}, an initialization of the hyperplanes $\mathcal{H}_{init} = \{(\textbf{n}_{k}, d_{k})\}_{k=1}^m$ is required. However, the true number of hyperplanes $m$ (model order) is typically unknown in unsupervised settings. To address this, we adopt an incremental model selection strategy: we iterate $m$ from $1$ to a theoretical maximum of $\lfloor \frac{n}{\textbf{dim}} \rfloor$. For each candidate $m$, we randomly sample initial values on the manifold and minimize the objective function via Sec.\ref{manifold_optimization}.


As the manifold $\mathcal{S}^{\textbf{dim}-1}$ is a nonlinear space embedded in an Euclidean space $\mathbb{R}^{\textbf{dim}}$, achieving equidistant sampling on this manifold is a challenging task. In this part, we introduce trigonometry to represent the normal vector $\textbf{n}$ and uniquely represent the hyperplane $h$. Although this method cannot achieve uniform sampling in higher dimensions, it can improve sampling efficiency by reducing the state space. Note that this method can achieve equidistant sampling in 2D space. And in 3D space, we also use an approximate approach to make the sampling relatively equidistant, which is discussed in detail in Appendix~\ref{appdend_sampling}.

A point $\textbf{n}$ on manifold $\mathcal{S}^{\textbf{dim}-1}$ can be described by a series angles $(\theta_1, \theta_2, ..., \theta_{\textbf{dim}-1})$ through:
\begin{equation}
\label{eq:coord_transfer}
\begin{gathered}
\left\{\begin{array}{l}
n_\textbf{dim} = \sin{\theta_{\textbf{dim}-1}} \\
n_{\textbf{dim}-1} = \cos{\theta_{\textbf{dim}-1}}\cdot\sin{\theta_{\textbf{dim}-2}}\\\vdots\\
n_k = (\prod_{i=k}^{\textbf{dim}-1}{\cos{\theta_{i}}})\cdot\sin{\theta_{k-1}}\\\vdots\\
n_2 = (\prod_{i=2}^{\textbf{dim}-1}{\cos{\theta_{i}}})\cdot\sin{\theta_{1}}\\
n_1 = \prod_{i=1}^{\textbf{dim}-1}{\cos{\theta_{i}}}
\end{array}\right.
\end{gathered}
\end{equation} where $n_k$ is the $k$-th element of $\textbf{n}$, and this equation is detailed in the Appendix~\ref{appdend_transfer}. If angles are known, the normal vector can be computed from $n_{\textbf{dim}}$ to $n_1$. Moreover, eq.\ref{eq:coord_transfer} can transform the sampling space form $\mathcal{S}$ to linear space $\mathbb{P}\times\mathbb{Q}^{\textbf{dim}-2}$, where $\times$ denotes the Cartesian product of two sets, $\mathbb{P}$ and $\mathbb{Q}$ are two intervals $[-\pi, \pi]$ and $[-\frac{\pi}{2}, \frac{\pi}{2}]$ respectively. Note that $\theta_1 \in \mathbb{P}, \theta_k\in\mathbb{Q}, \forall k \ge 2$.

Once the normal vector $\textbf{n}$ and distance $d \in \mathbb{R}_{+}$ are determined, we can get the initial hyperplane $h$. We denote a point $\hbar$ on hyperplane to represent $h$:
\begin{equation} \label{eq:hbar}
\hbar = d\textbf{n}
\end{equation} For a hyperplane in Euclidean space, we can easily get the point $\hbar$ is the point on the hyperplane that is closest to the origin, i.e.
\begin{equation}
\hbar = \arg\min_{x \in h} \lVert x \rVert_2
\end{equation}
And we can obviously get $\forall x \ne \textbf{0}, \exists ! h: \hbar = x$ (Appendix.~\ref{proof_hbar}). i.e. a point $\hbar$ can represent the unique hyperplane $h$. The case $d=0$ (i.e. $x = \textbf{0}$) can be ignored through changing the origin of the Cartesian coordinate system, which is easily achieved if the number of points is not infinite.

\section{Initial Value for Optimization} \label{initial}
Since it has a higher computational cost to solve the problem eq.\ref{eq:optimization_problem} using Sec.\ref{method} and Sec.\ref{sampling} with iteratively increasing $m$ from $1$ to $\frac{n}{\textbf{dim}}$, here we propose an approach to determine the number of hyperplanes and their initial values before deploying method in Sec.\ref{method} for computational efficiency.

The procedure for getting the initial values is shown in Algo.\ref{alg:initial_values}. The core of this algorithm is to iteratively find the hyperplane with the highest probability over the points in $\mathcal{X}$. Therefore, it can generate a number of hyperplanes $\textbf{card}(\mathcal{H})$ and their initial value. The function calls to \texttt{BestHyperplane} in the loop (lines 6-8) are independent. Consequently, we can execute these statements through parallel computing to enhance algorithmic efficiency.

\begin{algorithm}[tb]
   \caption{Initial values of hyperplanes}   \label{alg:initial_values}
    \begin{algorithmic}[1]
   \STATE {\bfseries Input:} a set of points $\mathcal{X} = \{x_1, \cdots, x_n\}$.
   \STATE {\bfseries Output:}  a set of initial hyperplanes $\mathcal{H} = \{h_1, \cdots, h_m\}$.

    \STATE $\mathcal{H} = \varnothing$
   \REPEAT

    \STATE Get a set of initial normal vector $\mathcal{N} = \{\textbf{n}_1, \cdots, \textbf{n}_M\}$ by Sec.\ref{sampling}. The series of angles $(\theta_1, \cdots, \theta_{\textbf{dim}})$ are picked evenly.
    \FOR{$\textbf{n}$ {\bfseries in} $\mathcal{N}$}

    \STATE $\mathcal{X}_p,d \gets \texttt{BestHyperplane}(\textbf{n}, \mathcal{X})$
    
    \ENDFOR

    \STATE Update $\mathcal{H} = \mathcal{H} \cup h$, $h = (\textbf{n}, d)$ is the hyperplane in loop (line 5) with highest number of points $\textbf{card}(\mathcal{X}_p)$.

    \STATE Update $\mathcal{X} = \mathcal{X} \diagdown \mathcal{X}_p$

   \UNTIL{$\textbf{card}(\mathcal{X}) \le thres$}
   
    \STATE {\bfseries return} $\mathcal{H}$
\end{algorithmic}
\end{algorithm}

Algo.\ref{alg:best_hyperplane} demonstrates the function \texttt{BestHyperplane} while the normal vector $\textbf{n}$ is determined. After all points are projected to the vector $\textbf{n}$ and sorted, we can get the maximum and minimum distances $ u$ and $ l$, respectively. In line 9, we uniformly sample $W$ windows. Then we will obtain the distance parameter of the best hyperplane based on the number of points contained within those windows, as shown in lines 10-15. Similar to Algo.\ref{alg:initial_values}, the computational efficiency of the Algo.\ref{alg:best_hyperplane} can be improved through parallel computing for lines 10-14.

\begin{algorithm}[tb]
   \caption{~\texttt{BestHyperplane}}   \label{alg:best_hyperplane}
    \begin{algorithmic}[1]
   \STATE {\bfseries Input:} a set of points $\mathcal{X}$, a normal vector $\textbf{n}$, the width of the window $W$.
   \STATE {\bfseries Output:}  a set of points $\mathcal{X}_p$, a distance value of the hyperplane $d$.

   \FOR{$x_i$ {\bfseries in} $\mathcal{X}$}
   \STATE $d_i = x_i \cdot \textbf{n}$
   \ENDFOR

    \STATE Sorting $\mathcal{X}$ according to $d_i$ from smallest to largest.
    \STATE $l \gets$ the smallest $d_i$, and $d_i > 0$.
    \STATE $u \gets$ the largest $d_i$
    \STATE Computing the number of the windows $\textbf{num} = \lceil \frac{2(u-l)}{W} \rceil$
    \FOR{$k$ {\bfseries from} $1$ {\bfseries to} $\textbf{num}$}
    \STATE the lower bound of window $W_l = l + (k-1)\frac{W}{2}$
    \STATE the upper bound of window $W_u = l + (k+1)\frac{W}{2}$
    \STATE Recording the points contained by the window: $\mathcal{X}_p^k = \{x_i | W_l \le d_i \le W_u, ~x_i \in \mathcal{X}\}$
    
    \ENDFOR
    
    \STATE Getting set of points $\mathcal{X}_p = \mathcal{X}_p^k$ with maximum $\textbf{card}(\mathcal{X}_p^k)$, and distance value $d = l + k\frac{W}{2}$ 

    \STATE {\bfseries return} $\mathcal{X}_p,~~d$
\end{algorithmic}
\end{algorithm}

\section{Experiments} \label{experiments}

We conduct comprehensive experiments to validate the effectiveness of our proposed framework. Our evaluation strategy consists of two main parts: comparative analysis against state-of-the-art baselines and ablation studies to assess the contribution of specific algorithmic components detailed in Sec.\ref{method}, \ref{sampling}, and \ref{initial}. In this section, we first define the evaluation metrics. Details regarding dataset generation and specifications are provided in Appendix \ref{data_generation}. Moreover, to avoid a zero denominator in the weighting model, we introduce a small constant $\varepsilon = 10^{-8}$.

 \subsection{Evaluation Metrics}



Given that the true model order (number of hyperplanes) is unknown in our problem setting, we employ a multi-faceted evaluation strategy using the following three metrics:

\begin{itemize}
 \vspace{-0.5em}
\setlength{\itemsep}{0.2em}    
\setlength{\parskip}{0.0em}  
    \item \textbf{Hyperplanes Number (HN):} The estimated number of hyperplanes $m$. A value closer to the ground truth $M$ indicates better model order selection.
    
    \item \textbf{Total Cost (TC):} The final value of the objective function (Eq.\ref{eq:original_obj}), representing the aggregate Euclidean distance from all points to their assigned hyperplanes. Lower values indicate a better geometric fit.
    
    \item \textbf{Hyperplanes Error (HE):} We utilize the feature vector $\hbar$ (defined in Eq.\ref{eq:hbar}) to quantify parameter estimation accuracy. Let $\hbar_i$ denote the $i$-th estimated hyperplane and $\hat{\hbar}_j$ denote the $j$-th ground truth hyperplane. The error is defined as the cumulative distance from each estimated hyperplane to its nearest ground truth counterpart:
    \begin{equation}
        HE = \sum_{i=1}^m \min_{j \in \{1,\dots,M\}} \lVert \hbar_i - \hat{\hbar}_j \rVert_2
    \end{equation}
    where $m$ and $M$ are the number of estimated and ground truth hyperplanes, respectively. Lower values are preferred.
\vspace{-0.5em}
\end{itemize}

It is important to acknowledge the challenge of \textbf{finite sample bias}: as noted by \cite{kulinski2023}, even when points are generated from a known manifold, parameters estimated from a finite sample set will inherently deviate from the ground truth due to sampling noise. Consequently, perfect alignment (zero HE) is theoretically unattainable. Furthermore, there exists a trade-off between these metrics: an over-estimated model order ($m > M$) typically reduces the Total Cost (TC) but may inflate the cumulative Hyperplane Error (HE) due to the summation of redundant terms. Conversely, under-estimation ($m < M$) minimizes the HE summation terms but leads to a significant increase in TC due to poor data fitting.

 \subsection{Ablation Experiments} \label{sec:ablation_experiments}
 In this section, each data instance consists of 120 points in a 2D space of size $10m \times 10m$, with a noise level of $\delta = 0.1m$. We randomly generated 20 data instances for each case. We conducted 10 independent trials for each of the six methods, as reported in Tab.\ref{tab:ablation_result}.

To better reflect the algorithm's performance, the number of hyperplanes corresponding to the results in the table is kept consistent with the ground truth for methods that fail to recover the model order. For method mentioned in Sec.\ref{initial}, we set the number of $\textbf{n}$ in Algo.\ref{alg:initial_values}(line 6) is 180. Those normal vectors are uniformly sampled through Appendix.\ref{appdend_sampling}. And the windows $W$ in algo.\ref{alg:best_hyperplane} is set $W = 4\delta$. For the probabilistic weighting in Phase I, we employ the inverse-square kernel (as defined in Algo.\ref{alg:manifold_optimization}) for all experiments, as our preliminary studies (see Appendix \ref{weight_test}) indicated its superiority over Gaussian and linear kernels.
 
\begin{table*}[ht]
\vskip -0.1in
\caption{Result of the ablation experiments.}
\label{tab:ablation_result}
\vskip 0.05in
\begin{center}
\begin{small}
\begin{sc}
\begin{tabular}{cc|ccc|ccc} 
\toprule
\multirow{2.5}{*}{\makecell[c]{Number of\\ Hyperplanes}} & \multirow{2.5}{*}{Metrics}& \multicolumn{6}{c}{Algorithm Settings} \\
\cmidrule(lr){3-8} 
~ & ~ & \multicolumn{3}{c}{Random Sampling} & \multicolumn{3}{c}{Initial Value(Sec.\ref{initial})} \\
\cmidrule(lr){3-5} \cmidrule(lr){6-8} 
~ & ~ & Soft (I) & Hard (II) & Full & Soft (I) & Hard (II) & Full \\
\midrule

\multirow{3}{*}{2} 
   & HN      & - & - & - & 2.0 & 2.0 & 2.0 \\
   & TC     & 24.207 & 31.012 & \textit{24.866}  & 5.709 & 6.459 & \textbf{5.706} \\
   & HE     & 0.981 & 0.961 & \textit{0.900}   & 0.030 & 0.097 & 0.029 \\  
\midrule
\multirow{3}{*}{3} 
   & HN      & - & - & - & 3.0 & 3.0 & 3.0 \\
   & TC     & 32.926 & 63.886 & \textit{39.640} & 5.720 & 6.623 & \textbf{5.712} \\
   & HE     & 2.662 & 4.203 & \textit{3.838} & 0.054 & 0.156 & 0.053 \\ 
\midrule
\multirow{3}{*}{4} 
   & HN      & - & - & - & 4.0 & 4.0 & 4.0 \\
   & TC     & 30.307 & 55.677 & 27.568 & 5.589 & 6.451 & \textbf{5.581}\\
   & HE     & 3.563 & 5.790 & 3.511 & 0.070 & 0.188 & 0.066\\ 
\midrule
\multirow{3}{*}{5} 
   & HN      & - & - & - & 5.0 & 5.0 & 5.0 \\
   & TC     & 27.587 & 48.359 & 26.539 & 5.671 & 6.677 & \textbf{5.653}\\
   & HE     & 4.294 & 6.475 & 4.201 & 0.099 & 0.279 & 0.102\\ 
\midrule
\multirow{3}{*}{6} 
   & HN      & - & - & - & 5.8 & 5.8 & 5.8 \\
   & TC     & 28.275 & 46.190 & 26.838 & 8.838 & 9.151 & \textbf{8.186}\\
   & HE     & 5.771 & 7.735 & 5.634 & 0.680 & 0.673 & 0.611 \\ 

\bottomrule
\end{tabular}
\end{sc}
\end{small}
\end{center}
\vskip -0.2in
\end{table*}

In ablation experiments, the Full Pipeline (Phase I + II) consistently outperforms using Phase I or Phase II alone. Meanwhile, the more hyperplanes there are, the greater the difference in performance. This aligns with our design philosophy: Phase I (Soft Optimization) effectively navigates the complex landscape to avoid local optima, while Phase II (Hard Refinement) ensures the solution converges to the precise geometric truth.

For initial value, our methods (Sec.\ref{initial}) can help manifold optimization algorithms achieve the global optimum, compared to random sampling. However, when the number of hyperplanes increases and the points are mixed, this method also yields a number of hyperplanes different from the ground truth, i.e. finding 5 hyperplanes while true number is 6 in some data ($HN = 5.8$).

In summary, Sec.\ref{initial} can find the available initial hyperplanes, and manifold optimization can solve the whole problem. 

\subsection{Compared with Baselines}
As the problem eq.\ref{eq:original_obj} can be regarded as unsupervised learning or clustering, we compared our methods with the following classic algorithms: Agglomerative Clustering\cite{Johnson1967agg}, K-Means\cite{arthur2007kmeans}, DBSCAN\cite{ester1996density}, OPTICS\cite{ankerst1999optics}, GMM\cite{dempster1977maximum}, and RANSAC\cite{fischler1981ransac}. We have improved those methods to adapt to the problem, computing the distance between points and clusters using the distance between points and the hyperplanes.

Moreover, we also compare our methods with two widely known baselines, PARSAC\cite{kluger2024parsacacceleratingrobustmultimodel} and SupeRANSAC\cite{barath2025superansacransacrule}. As these two algorithms are mainly for image processing, we implement simplified versions of PARSAC and SupeRANSAC adapted for the problem hyperplane fitting, preserving their core algorithmic principles (parallel hypothesis generation, soft inlier weighting, greedy selection for PARSAC; PROSAC sampling, MAGSAC scoring for SupeRANSAC) while replacing domain-specific components with geometric alternatives suitable for our task. There are four variants of these baselines:

 \begin{itemize}
 \vspace{-0.5em}
 \setlength{\itemsep}{0.2em}    
\setlength{\parskip}{0.0em}  
\item PARSAC known: We set the number of hyperplanes as the ground truth to test the PARSAC, i.e., the number of hyperplanes is known.
\item PARSAC unknown: The algorithm does not know the number of hyperplanes, and the method is terminated when the remaining points are few.
\item SupeRANSAC known: The method knows the ground truth of hyperplanes' number.
\item SupeRANSAC unknown: This is similar to PARSAC unknown.
 \vspace{-1.em}
 \end{itemize}

\begin{table*}[t]
\vskip -0.1in
\caption{Results of comparison. "-" means the number of hyperplanes is known.}
\label{tab:compare_result}
\vskip 0.05in
\begin{center}
\begin{small}
\begin{sc}
\begin{tabular}{l|ccc|ccc|ccc|c}
\toprule
\multirow{2}{*}{Methods} &
\multicolumn{3}{c|}{3 Hyperplanes} & \multicolumn{3}{c|}{4 Hyperplanes} & \multicolumn{3}{c|}{5 Hyperplanes} & \multirow{2}{*}{Avg} \\
\cmidrule{2-10}
~ & HN & TC & HE & HN & TC & HE & HN & TC & HE & Time (s) \\
\midrule

 AggCluster        & - & 47.188 & 2.045 & - & 43.531 & 2.752 & - & 46.248 & 4.249 & 0.002 \\
 K-Means           & - & 62.638 & 4.898 & - & 55.650 & 7.085 & - & 54.433 & 8.619 & 0.012 \\
 DBSCAN            & 4.15 & 85.481 & 5.471 & 3.5 & 113.258 & 5.294 & 2.95 & 142.709 & 4.035 & 0.001 \\
 OPTICS            & 6.75 & 20.760 & 5.146 & 6.5 & 31.820 & 6.400 & 6.4 & 39.935 & 5.668 & 0.032 \\
 GMM+EM            & - & 58.260 & 3.890 & - & 44.475 & 3.819 & - & 39.169 & 4.861 & 0.383 \\
 RANSAC            & - & 302.270 & 8.272 & - & 145.392 & 8.735 & - & 95.745 & 9.036 & 0.027 \\
 PARSAC known      & - & 17.347 & 0.202 & - & 16.634 & 0.290 & - & 15.790 & 0.447 & 0.021 \\
 PARSAC unknown    & 6.85 & 8.210 & 2.403 & 7.20 & 9.643 & 1.732 & 7.15 & 11.658 & 1.382 & 0.091 \\
 SupeRANSAC known  & - & 25.028 & 0.534 & - & 18.569 & 0.674 & - & 19.479 & 0.941 & 0.032 \\
 SupeRANSAC unknown & 6.5 & 7.512 & 1.427 & 8.10 & 8.277 & 1.978 & 9.3 & 8.472 & 2.678 & 0.093 \\
 
\midrule
 Random + Full Pipeline     & - & 36.445 & 2.752 & - & 42.552 & 5.498 & - & 36.983 & 6.221 & 0.423 \\
 Sec.\ref{initial} + Full Pipeline & 3.0 & \textbf{16.681} & 0.135 & 4.0 & \textbf{16.590} & 0.241 & 5.1 & \textbf{15.727} & 0.616 & 0.357 \\
\bottomrule
\end{tabular}
\end{sc}
\end{small}
\end{center}
\vskip -0.2in
\end{table*}

\begin{figure}[h]
\vskip -0.05in
\begin{center}
\includegraphics[width=0.43\textwidth]{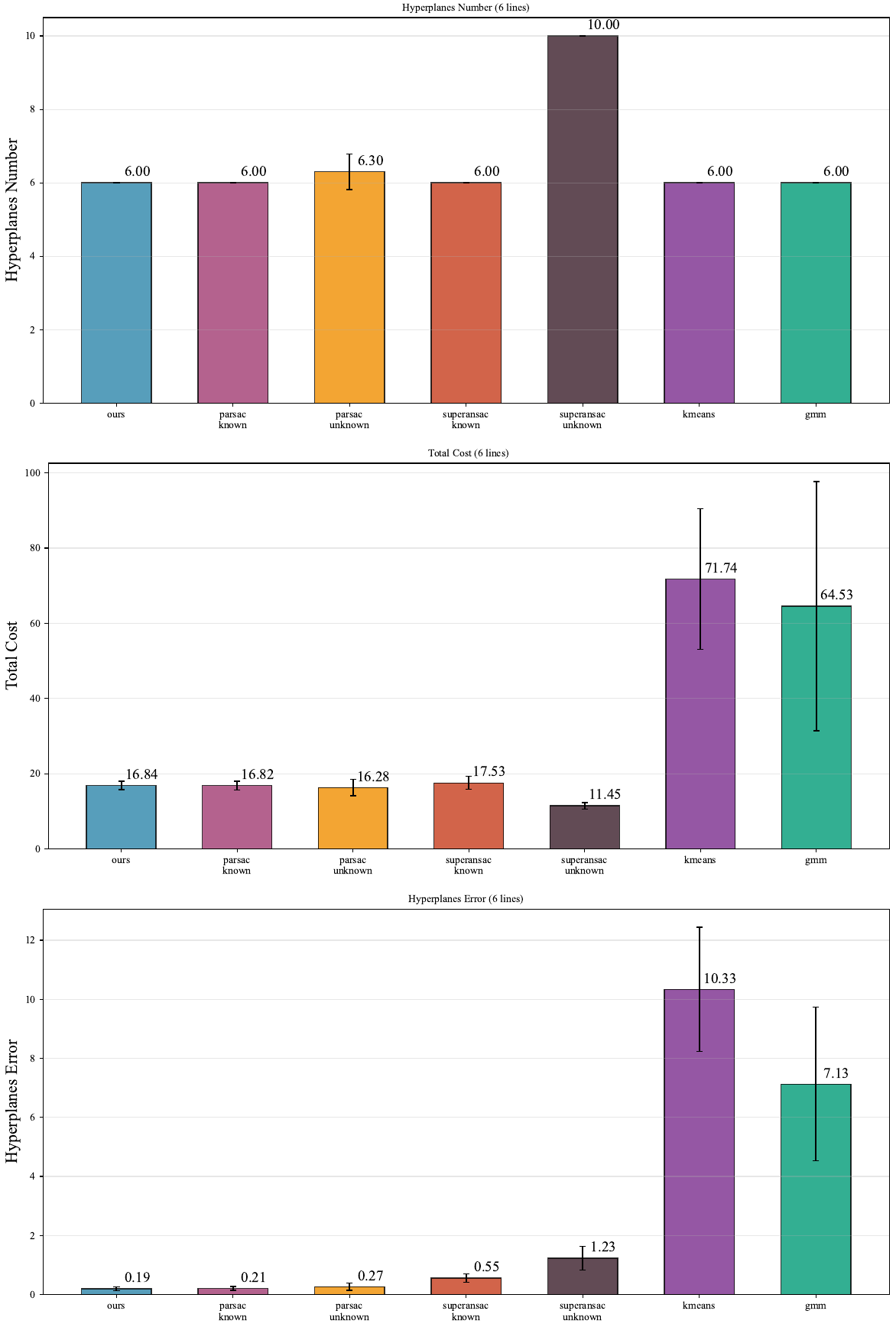}
\caption{Some results of comparison in six hyperplanes. Ours means initial value(Sec.\ref{initial}) + Full Pipeline.}
\label{fig:visual_compared_six}
\end{center}
\vskip -0.1in
\end{figure}

In this comparative study, we utilize synthetic 2D datasets bounded within a $10m \times 10m$ region, with each instance containing $N=120$ points. The noise level is set to $\delta = 0.3m$. To ensure statistical reliability, we generate 20 independent data instances for each ground truth configuration (number of hyperplanes). We report the average performance over 10 independent trials per method, as summarized in Tab.\ref{tab:compare_result} and Fig.\ref{fig:visual_compared_six}. For our framework, the window parameter is fixed at $W = 2\delta$.



As demonstrated in Tab.\ref{tab:compare_result}, our \textbf{Full Pipeline} (Sec.\ref{initial} + Phase I + II) achieves the best overall performance compared to all baselines. Although the estimated model order ($HN = 5.1$) deviates slightly from the ground truth ($M=5$), our method yields the lowest Total Cost and Hyperplane Error, indicating superior geometric accuracy.

The performance gap can be attributed to the inherent geometry of the problem. When formulated as a traditional clustering task, algorithms struggle with "intersection ambiguity": unlike compact clusters, hyperplanes intersect in Euclidean space. Points near intersections are geometrically ambiguous, causing classic algorithms (e.g., K-Means, DBSCAN) to misclassify them, resulting in poor structural recovery. Regarding RANSAC-based methods (PARSAC and SupeRANSAC), they exhibit competitive performance only when the ground truth model order is explicitly provided ("known" setting). In contrast, their performance degrades significantly in the unsupervised setting ("unknown"). In these cases, they tend to severely overestimate the number of hyperplanes, leading to a failure in capturing the true structure.

Fig.\ref{fig:visual_compared_six} demonstrates the results where there are 6 hyperplanes in data instance. While our method (Full Pipeline) is slightly inferior to PARSAC known in terms of the Total Cost, it not only identifies the exact number of hyperplanes accurately but also achieves the lowest Hyperplane Error.

For newer RANSAC-based methods, when the number of hyperplanes is unknown, they tend to overestimate the number of hyperplanes, which leads to increase in the Hyperplane Error and decrease in the Total Cost. For other classic methods, they are hardly able to identify the affiliation of each point to the corresponding hyperplanes when hyperplanes intersect, which results in their poor performance. These performances are similar to that in Tab.\ref{tab:compare_result}.

\section{Conclusion} \label{conclusion}


This work studies the problem of fitting points with an unknown number of hyperplanes, which is inherently non-convex and non-differentiable. To circumvent the non-differentiability, we reformulate the task within a clustering framework. Subsequently, we develop a solution based on manifold optimization. Specifically, we introduce an initialization strategy that enables searching and sampling in non-convex space and design a Riemannian optimization algorithm to solve the solution, effectively handling the geometric constraints. Experimental results demonstrate that our proposed method successfully solves the problem and significantly outperforms existing baselines.

\clearpage

\bibliography{paper}
\bibliographystyle{icml2025}

\newpage
\appendix
\onecolumn

\section{Appendix Part I}
\subsection{Manifold Optimization} \label{pre_manifold}
Manifold optimization\cite{hu2019briefintroductionmanifoldoptimization, Boumal_2023, James2016pymanopt} is a methodology to solve the optimization problem with constraints. Specifically, it constructs a manifold to describe the constraints, thereby transforming the optimization problem into an unconstrained one on the manifold. The paradigm of the manifold optimization is
\begin{equation}
\begin{gathered}
\min f(x) \\
\begin{matrix}
\text { s. t. } & x \in \mathcal{M} 
\end{matrix}
\end{gathered}
\end{equation} where $\mathcal{M}$ is the manifold representing the constraints.



When employing standard numerical methods, $x$ may drift off the manifold $\mathcal{M}$, rendering the solution invalid with respect to the original constraints. To ensure the validity of the solution, manifold optimization introduces specific operations to guarantee $x \in \mathcal{M}$. These operations are detailed below:

\textbf{Tangent Space}: For any point $x \in \mathcal{M}$, the tangent space is a linear space represented by $\mathcal{T}_x\mathcal{M}$, comprising all tangent vectors at point $x$ on the manifold $\mathcal{M}$.

\textbf{Projection}: This operation projects a vector into the tangent space defined at point $x$, i.e. $\textbf{Proj}_x(\cdot) \in \mathcal{T}_x\mathcal{M}$. It satisfies $\Big(v - \textbf{Proj}_x(v)\Big) \perp \mathcal{T}_x\mathcal{M}.$

\textbf{Riemannian Gradient}: The Riemannian gradient of function $f(x)$ on point $x$ belong to the tangent space $\mathcal{T}_x\mathcal{M}$, which is labeled as $\textbf{grad}f(x)$. The direction of the Riemannian gradient is the direction that maximizes the directional derivative $df(x)$. To compute the Riemannian gradient of $f(x)$ at point $x$, we can compute the Euclidean gradient and then project it to the tangent space $\mathcal{T}_x\mathcal{M}$.

\textbf{Retraction}: Retraction is a function that maps a vector in tangent space into a point $y \in \mathcal{M}$. This operation is the key to guarantee $x \in \mathcal{M}$, as the computed value $\hat{x}$ according to the Riemannian gradient may not be on the manifold. Let $\mathcal{R}_x(v)$ denotes the retraction on point $x$, where $v \in \mathcal{T}_x\mathcal{M}$. It is noted that $\mathcal{R}_x(\textbf{0}) = x$.

In summary, updating $x$ via manifold optimization involves the following steps: 1) Computing the Euclidean gradient of the objective function. 2) Computing the Riemannian gradient based on the Euclidean gradient through projection. 3) Computing the new value $\hat{x}$ based on Riemannian gradient. 4) Retract $\hat{x}$ onto the manifold $\mathcal{M}$.

\subsection{Equidistant Sampling on $\mathcal{S}^{\textbf{dim}-1}$ with $\textbf{dim} = 2$ and $3$.} \label{appdend_sampling}

\begin{figure}[ht]
\vskip 0.2in
\begin{center}
\includegraphics[width=0.6\textwidth]{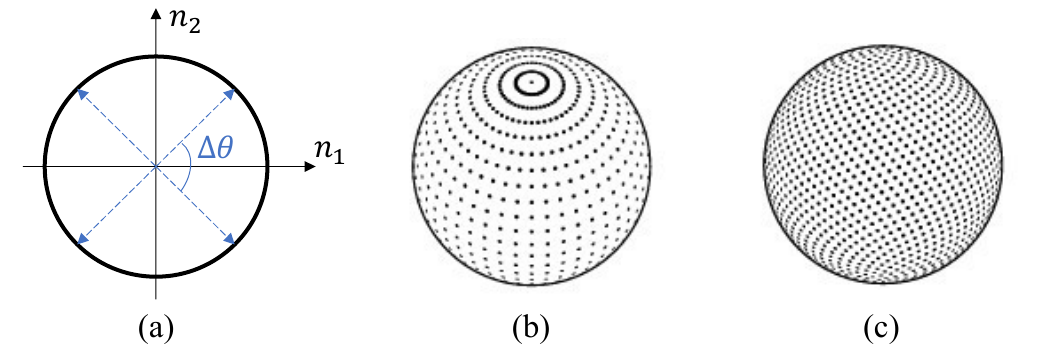}
\caption{Illustration of sampling on manifold $\mathcal{S}^1$ and $\mathcal{S}^2$. (a) shows equidistant sampling on $\mathcal{S}^1$. (b) and (c) are sampling on $\mathcal{S}^2$.}
\label{fig:uniformly_sampling}
\end{center}
\vskip -0.2in
\end{figure}

Normal vector $\textbf{n}$ can be represented by $(\cos\theta_1, \sin\theta_1)$ and $(\cos\theta_1\cos\theta_2, \sin\theta_1\cos\theta_2 ,\sin\theta_2)$ in space $\mathbb{R}^2$ and $\mathbb{R}^3$ respectively. For manifold $\mathcal{S}^1$ embedded in $\mathbb{R}^2$, we can easily achieve equidistant sampling of normal vector by sampling $\theta_1$ equidistantly in the linear space $\mathbb{P} = [-\pi, \pi]$, as shown in Fig.~\ref{fig:uniformly_sampling}(a). However, if sampling $\theta_1 \in \mathbb{P}$ and $\theta_2 \in \mathbb{Q} = [-\frac{\pi}{2}, \frac{\pi}{2}]$ equidistantly, we can not achieve evenly sampling, which is shown in Fig.~\ref{fig:uniformly_sampling}(b). The closer $\theta_2$ is to $-\frac{\pi}{2}$ or $\frac{\pi}{2}$, the higher the sampling density.

For sampling on the manifold $\mathcal{S}^2$, {\'A}lvaro~\cite{Alvaro2010} proposed a method to achieve approximately uniform sampling; the sampling result is shown in Fig.\ref{fig:uniformly_sampling}(c). 



\subsection{Mapping between normal vector \textbf{n} and series of angles $\theta$.} \label{appdend_transfer}

We define $P_i = (n_1, n_2, ..., n_i, 0, ..., 0) \in \mathbb{R}^{\textbf{dim}}$ is a point in \textbf{dim}-dimension Cartesian coordinate system. A point $P_{\textbf{dim}} = (n_1, n_2, ..., n_{\textbf{dim}})$ in \textbf{dim}-dimension Cartesian coordinate system can be described by $(1, \theta_1, \theta_2, ..., \theta_{\textbf{dim}-1})$ in D-sphere coordinate system. $\theta_i$ is the angle between two vector $\overrightarrow{OP_{i+1}}$ and $\overrightarrow{OP_{i}}$. The transformation of the point $P_\textbf{dim}$ in two coordinate system is eq.\ref{eq:coord_transfer} and
\begin{equation}
\theta_i = sign(n_{i+1})\cdot\arccos{\frac{\sqrt{n_1^2 + n_2^2 + ... + n_i^2}}{\sqrt{n_1^2 + n_2^2 + ... + n_{i+1}^2}}}
\end{equation} where $sign(\cdot)$ is the symbol function.

\begin{figure}[ht]
\vskip 0.2in
\begin{center}
\includegraphics[width=0.6\textwidth]{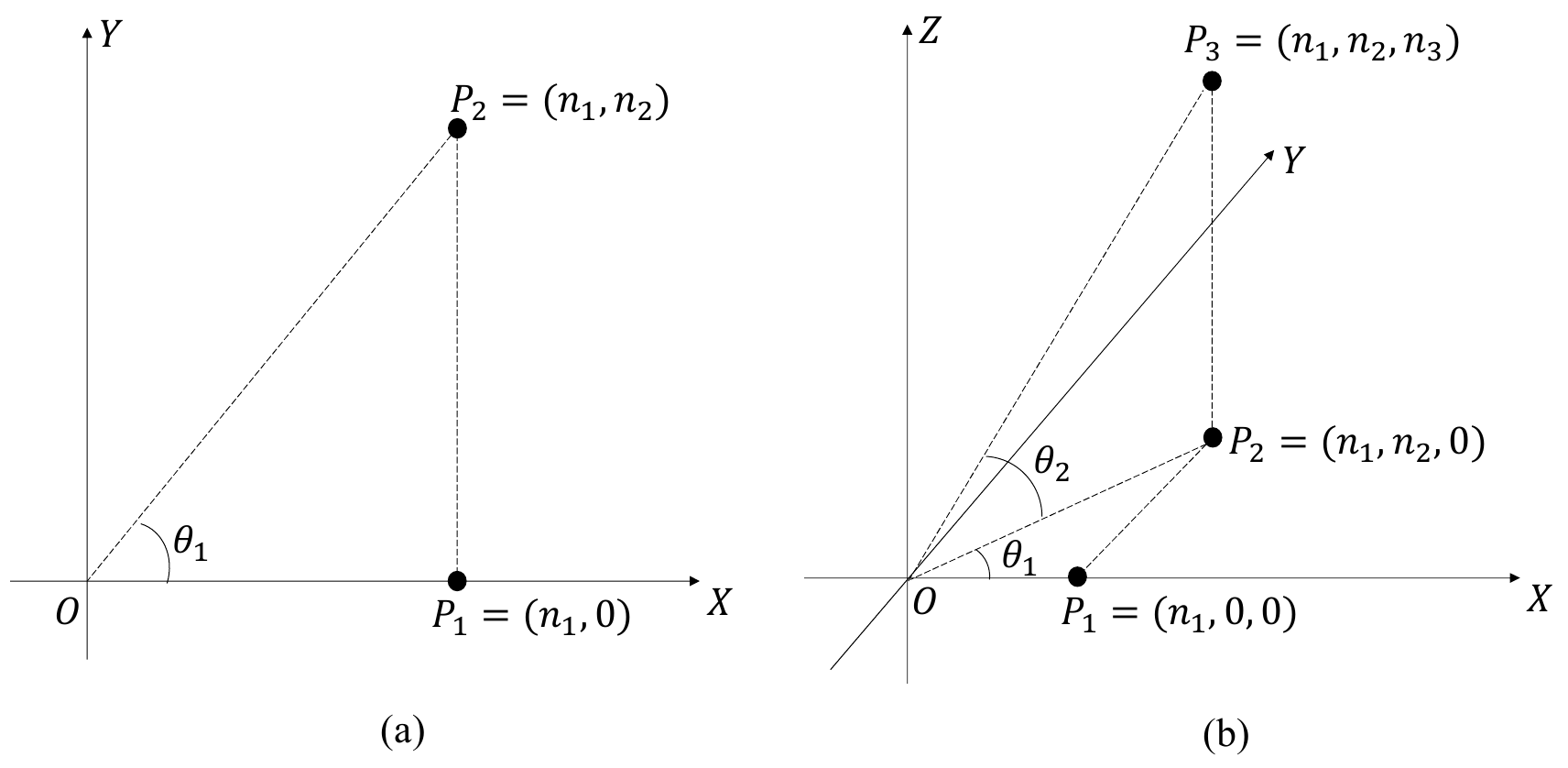}
\caption{Illustration of mapping. (a) shows the case in $\mathbb{R}^2$. (b) shows the case in $\mathbb{R}^3$.}
\label{fig:point_transfer}
\end{center}
\vskip -0.2in
\end{figure}

As $\mathbb{R}^i \subset \mathbb{R}^{i+1}$ we make assumption that the axis and origin of $i$-dimension Cartesian coordinate system are the same as that in the $(i+1)$-dimension Cartesian coordinate system. When transfer $\overrightarrow{OP_{i}} \in \mathbb{R}^i$ to $\overrightarrow{OP_{i}} \in \mathbb{R}^{i+1}$, the $\theta_i$ is the angle between two vectors $\overrightarrow{OP_{i}}, \overrightarrow{OP_{i+1}}$ both belong to $\mathbb{R}^{i+1}$. Therefore, by iteratively projecting vector $\textbf{n}$ from a high-dimensional space to a low-dimensional space, those $\theta$ can be computed. Examples in 2D and 3D space are shown in Fig.~\ref{fig:point_transfer}.

\subsection{The point $\hbar$ of hyperplane} \label{proof_hbar}


As mentioned before, a hyperplane $h \in \mathbb{R}^{\textbf{dim}}$ can be represented by a normal vector $\textbf{n} \in \mathcal{S}^{\textbf{dim}-1}$ and a distance parameter $d \in \mathbb{R}$, denoted as $h(\textbf{n}, d) = \{x | \textbf{n} \cdot x = d\}$. 
Here we define $\hbar$ as the point on the hyperplane $h$ closest to the origin (i.e., with the minimum Euclidean norm). We propose the Proposition: $\forall x \ne \textbf{0}, \exists ! h: \hbar = x$.

\begin{proof}
Finding $\hbar$ is equivalent to solving the following constrained optimization problem:
\begin{equation}
\begin{aligned}
\min_{x} \quad & \frac{1}{2} \lVert x \rVert_2^2 \\
\textrm{s.t.} \quad & \textbf{n}^T x = d
\end{aligned}
\end{equation}
This is a convex optimization problem. We apply the method of Lagrange multipliers. The Lagrangian function is:
\begin{equation}
\mathcal{L}(x, \lambda) = \frac{1}{2} x^T x - \lambda (\textbf{n}^T x - d)
\end{equation}
Taking the derivative with respect to $x$ and setting it to zero (KKT stationarity condition):
\begin{equation}
\nabla_x \mathcal{L} = x - \lambda \textbf{n} = 0 \implies x = \lambda \textbf{n}
\end{equation}
Substituting $x = \lambda \textbf{n}$ into the constraint $\textbf{n}^T x = d$:
\begin{equation}
\textbf{n}^T (\lambda \textbf{n}) = d \implies \lambda (\textbf{n}^T \textbf{n}) = d
\end{equation}
Since $\textbf{n} \in \mathcal{S}^{\textbf{dim}-1}$, we have $\textbf{n}^T \textbf{n} = \lVert \textbf{n} \rVert_2^2 = 1$. Therefore, we obtain $\lambda = d$.
Substituting $\lambda$ back into the expression for $x$, we derive the unique analytical solution:
\begin{equation}
\hbar = d\textbf{n}
\end{equation}
Thus, for any vector $x \ne \textbf{0}$, we can uniquely determine the hyperplane parameters by setting $d = \lVert x \rVert_2$ and $\textbf{n} = x / \lVert x \rVert_2$, which yields $\hbar = x$. This proves the existence and uniqueness of the representation.
\end{proof}

\subsection{Model-order Recovery} \label{show_random_sampling}
\begin{figure}[h]
\vskip 0.2in
\begin{center}
\includegraphics[width=0.6\textwidth]{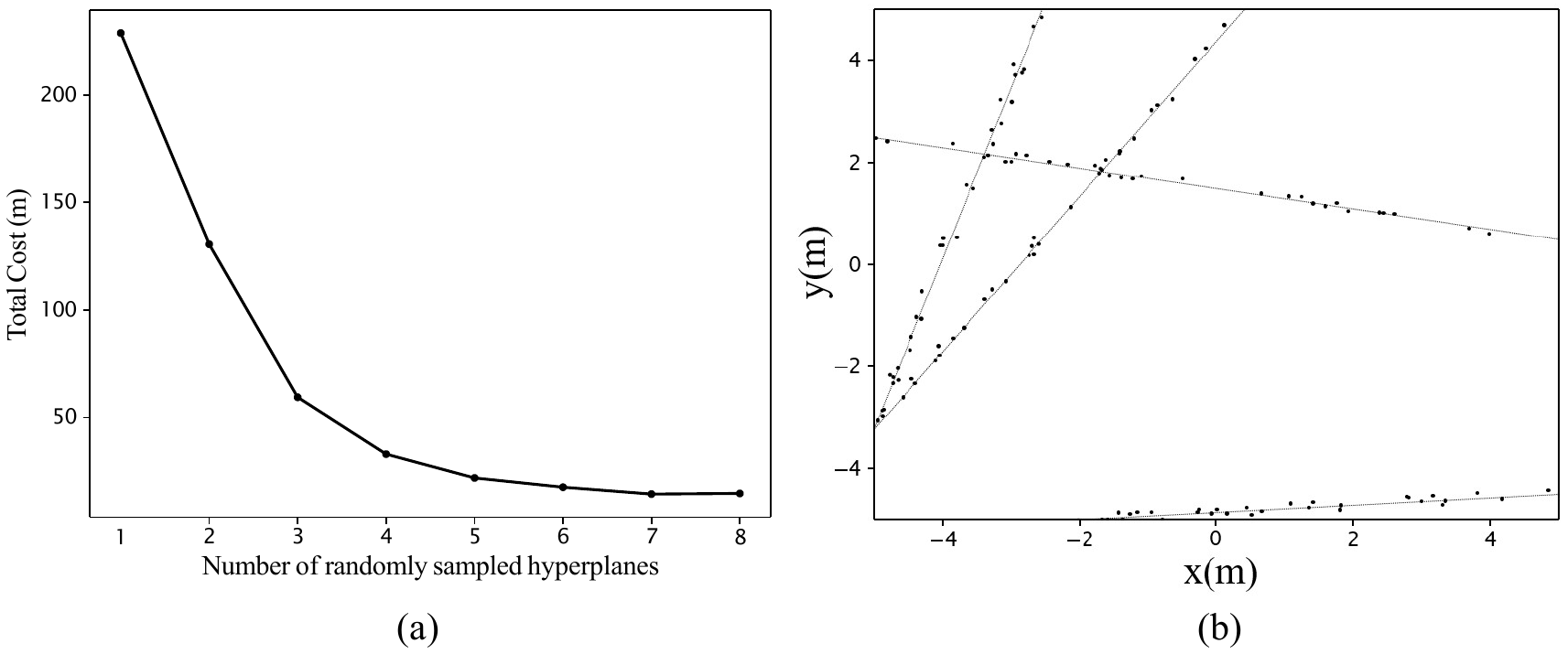}
\caption{Illustration of "elbow" shape example. (a) is the total cost when the number of sampled hyperplanes increases from 1 to 8. (b) is the data.}
\label{fig:elbow_data}
\end{center}
\vskip -0.2in
\end{figure}

Fig.\ref{fig:elbow_data} illustrates the "elbow" shape of random sampling. 

To the best of our knowledge, model order recovery remains a challenging problem\cite{pmlr-v27-luxburg12a, 10585323}, and to date, no existing work can guarantee accurate identification of the true model order. At present, the number of clusters can only be initialized at 1 and gradually increased, with the algorithm run sequentially. The appropriate model order is determined by evaluating performance and related metrics, for instance, the Bayesian Information Criterion (BIC). In our problem, the distance term in the objective function is defined using the L1-norm. Accordingly, the BIC criterion should be derived under the Laplace distribution assumption, i.e. $$2n \ln\!\left(\frac{2}{n} \sum r_i\right) + 2n + (m(dim+1)-1)\ln n,$$
where $n$ is the number of points, $m$ is the number of hyperplanes, $dim$ is the dimension, and $r_i$ is the distance between point $i$ and its closest hyperplane.

\subsection{Time Complexity}
Dimension and candidate model order can significantly influence the cost of initialization. We use a limited representation for the additional dimensions, and the time complexity with respect to dimension is $O(n)\cdot O(1)^{\mathrm{dim}-1}$. The time complexity with respect to the candidate model order is $O(n)$.

Dimension and candidate model order do not significantly influence the cost of soft optimization or hard refinement. During optimization, we use vectorized computation. At the same time, the optimization problems for multiple hyperplanes can be solved in parallel.

\subsection{Convergency} \label{convergency}
While clustering algorithms may yield satisfactory model order recovery performance when distinct boundaries exist between clusters, this is not the case for soft optimization scenarios. Specifically, when the parameter $\hbar$ of the hyperplanes is small, even human observers cannot reliably distinguish the actual number of hyperplanes (see Appendix. \ref{limitation} for details). For this reason, there is no convergence guarantee for Soft Optimization, and we only demonstrate the convergence of the hard refinement method in this work.

\subsubsection{Simplification}
In Hard Refinement (weights = 1), Eq. 12 can be simplified as
$$
\min_{n:\|n\|=1}\quad f(n) = \frac{1}{m}\sum_{i=1}^m \left( n^\top x_i - \frac{1}{m}\sum_{j=1}^m n^\top x_j \right)^2
$$
where $m$ is the number of points.
Let
$$
\bar{x} = \frac{1}{m}\sum_{i=1}^m x_i,
$$
then the objective function can be rewritten as
$$
\frac{1}{m}\sum_{i=1}^m \big(n^\top (x_i - \bar{x})\big)^2.
$$
Thus, the objective function simplifies to
$$
f(n) = n^\top S n,
$$
where
$$
S = \frac{1}{m}\sum_{i=1}^m (x_i - \bar{x})(x_i - \bar{x})^\top.
$$

\subsubsection{Solvability}
The sphere manifold is a bounded and closed set. By the Heine–Borel theorem, it is compact.

The objective function is a continuous quadratic function.

A continuous function on a compact set must attain its global minimum (Weierstrass extreme value theorem). Therefore, the optimization problem is solvable.

\subsubsection{Convergence Proof}
The Euclidean gradient of the objective function and manifold optimization are introduced in the paper. Combined with the retraction defined in the paper and the above simplification, we have

$$
n_{k+1} = \frac{n_k - \alpha \cdot \operatorname{grad} f(n_k)}{\|n_k - \alpha \cdot \operatorname{grad} f(n_k)\|}
= \frac{n_k - 2\alpha \big(S n_k - (n_k^\top S n_k) n_k\big)}{\|n_k - 2\alpha \big(S n_k - (n_k^\top S n_k) n_k\big)\|}.
$$

\textbf{Monotonic Decrease}

We can obtain
$$
\left. \frac{d}{d\alpha} f(\mathcal{R}_{n_k}(-\alpha \operatorname{grad} f(n_k))) \right|_{\alpha=0}
= -\|\operatorname{grad} f(n_k)\|^2 \le 0.
$$
If $n_k$ is not a critical point ($\operatorname{grad} f(n_k) \neq 0$), the derivative is negative, so there exists $\alpha > 0$ such that $f(n_{k+1}) < f(n_k)$.

\textbf{Boundedness of the Iterate Sequence}

All iterates $\{n_k\}_{k=0}^\infty$ lie on $S^{d-1}$, which is a compact set. Thus, the sequence $\{n_k\}$ is bounded, and by the Bolzano–Weierstrass theorem, it has a convergent subsequence $\{n_{k_j}\}$ such that $n_{k_j} \to n^* \in S^{d-1}$ as $j \to \infty$.

\textbf{Convergence}

We first show that the limit $n^*$ of the convergent subsequence is a critical point (i.e., $\operatorname{grad} f(n^*) = 0$). Since $f(n_k)$ is monotone decreasing and bounded below ($f(n) \ge 0$ because $S$ is positive semidefinite), $f(n_k) \to f(n^*)$ as $k \to \infty$. For the subsequence $\{n_{k_j}\}$, we have $f(n_{k_j + 1}) \to f(n^*)$.

Suppose $\operatorname{grad} f(n^*) \neq 0$. Then, for sufficiently large $j$, $\|\operatorname{grad} f(n_{k_j})\| \ge \epsilon > 0$ (by continuity of $\operatorname{grad} f$), and there exists $\alpha > 0$ such that $f(n_{k_j + 1}) \le f(n_{k_j}) - \delta$ for some $\delta > 0$. This contradicts $f(n_{k_j + 1}) \to f(n^*)$ and $f(n_{k_j}) \to f(n^*)$, so $\operatorname{grad} f(n^*) = 0$.

Next, we show that critical points of $f$ on $S^{d-1}$ are global optima. If $\operatorname{grad} f(n^*) = 0$, then
$$
2S n^* - 2(n^{*^\top} S n^*) n^* = 0,
$$
which implies
$$
S n^* = (n^{*^\top} S n^*) n^*.
$$
Thus, $n^*$ is an eigenvector of $S$, and $f(n^*) = n^{*^\top} S n^*$ is the corresponding eigenvalue. Since we are minimizing $f(n)$, $n^*$ must be an eigenvector corresponding to the minimum eigenvalue $\lambda_{\min}(S)$, i.e., a global optimum.

Finally, the entire sequence $\{n_k\}$ converges to a global optimum. Suppose there exists another subsequence converging to a different limit $\tilde{n}$. Then $\tilde{n}$ is also a global optimum, and $f(n^*) = f(\tilde{n}) = \lambda_{\min}(S)$. For the monotone sequence $f(n_k)$, this implies $n_k \to n^*$ (or $\tilde{n}$, which is also a global optimum), so the entire sequence converges to a global optimum of the problem.

\textbf{Conclusion}
Therefore, since manifold optimization can reach the global optimum of Eq. 12 for a single hyperplane, the proposed method can reach a local optimum for locating multiple hyperplanes.

\subsection{Limitation} \label{limitation}
The experimental results indicate the following issues in solving this problem. First, the distribution of data points affects the algo.\ref{alg:manifold_optimization}. Second, the initial hyperplanes affect the algorithm's ability to find the global optimum. The reasons for this are mentioned below.

\begin{figure}[b]
\vskip 0.2in
\begin{center}
\includegraphics[width=0.4\textwidth]{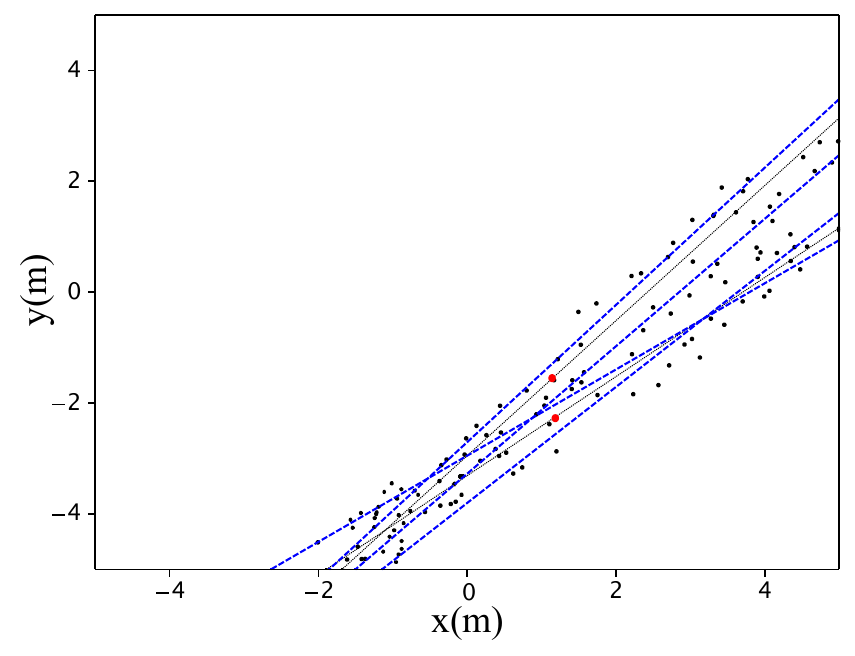}
\caption{Illustration of examples when $\hbar$ of two hyperplanes is close. The black points are data. The light black dotted lines are ground truth hyperplanes, while the dotted blue lines are the result. The red points are $\hbar$ of two hyperplanes.}
\label{fig:limitation}
\end{center}
\vskip -0.2in
\end{figure}

Since the number of hyperplanes is unknown, using this algorithm to solve the problem (eq.\ref{eq:original_obj}) may result in a number of hyperplanes different from the ground truth, making it impossible to obtain the correct solution. Specifically, when the features $\hbar$ of different hyperplanes are close, the data generated by these hyperplanes can become mixed together, causing the algorithm to misjudge the number of hyperplanes. Fig.\ref{fig:limitation} shows the example where $\hbar$ of two hyperplanes is close.

To address those limitations, the number of points on the hyperplane found by the algorithm can be limited, i.e., we can set a range for $\textbf{card}(\mathcal{X}^k_p)$ in line 15 of algo.\ref{alg:best_hyperplane}. And for data, we can increase the distance between $\hbar$ of any two ground truth hyperplanes, i.e.
$$\forall i, j, \quad  \lvert \hbar_i - \hbar_j \rvert_2 \gg \delta_i + \delta_j$$

\clearpage

\section{Appendix Part II}
\subsection{Data Generation} \label{data_generation}



To simulate the real-world scenario where the model order is unknown, we construct synthetic datasets containing varying numbers of ground truth hyperplanes, denoted as $M$. For the primary experiments, we maintain a fixed total sample size $N$ across all instances to ensure fair comparability.

The data generation process for a specific hyperplane $h(\bm{a}, b)$ is controlled by a noise parameter $\delta$. Points $x$ are randomly sampled to satisfy the bounded constraint:
\begin{equation}
    -\delta \le \bm{a}^Tx + b \le \delta
\end{equation}
This effectively creates a "slab" of width $2\delta$ around each hyperplane. For the standard dataset, we employ a \textbf{balanced distribution strategy}: given $M$ hyperplanes and a total capacity $N$, each hyperplane generates exactly $\frac{N}{M}$ points.


\subsection{Higher-dimensional Experiments}
We test our algorithm (initial value + soft + hard) on 4D and 5D synthetic data; we will add these results to the Appendix. The parameters in these experiments are the same as those in Table 1. Here, we only report the results for 3, 4, and 5 hyperplanes (Tab.\ref{tab:high_dim}). Increasing the dimension has a significant impact on the computation time of the initialization.

\begin{table}[h]
\label{tab:high_dim}
\caption{Experiments in 4D and 5D. 120 points and $\delta=0.1m$}
\begin{center}
\begin{tabular}{lccccccccc}
\toprule
 & \multicolumn{3}{c}{3 hyperplanes} & \multicolumn{3}{c}{4 hyperplanes} & \multicolumn{3}{c}{5 hyperplanes}\\
Dimension & HN & TC & HE & HN & TC & HE & HN & TC & HE\\
\midrule
4D & 3.0 & 4.976 & 0.095 & 4.0 & 5.263 & 0.116 & 5.0 & 5.386 & 0.144\\
5D & 3.0 & 5.428 & 0.111 & 4.0 & 5.393 & 0.195 & 5.0 & 5.117 & 0.189\\
\bottomrule
\end{tabular}
\end{center}
\end{table}

\subsection{Weight Matrix} \label{weight_test}
\begin{figure}[ht]
\vskip 0.2in
\begin{center}
\includegraphics[width=0.9\textwidth]{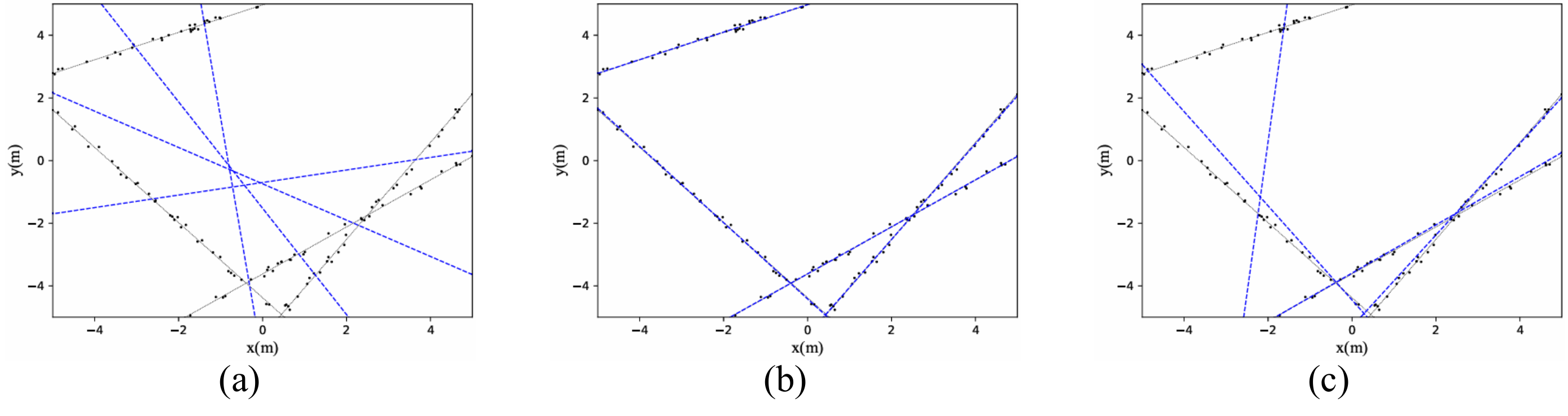}
\caption{Illustration of examples in 2D space with different computation methods of $R$. The dotted blue lines are found hyperplanes. (a) is $\frac{1}{D}$, (b) is $\frac{1}{D^2}$, and (c) is $e^{-D}$.}
\label{fig:weight_matrix_computation}
\end{center}
\vskip -0.2in
\end{figure}

For the weight matrix in line 8 of algo.\ref{alg:manifold_optimization} with $w_{ij} = \frac{R_{ij}}{\sum_{k=1}^m R_{ik}}$, we evaluated three kernel functions to model the probability density:
$$\frac{1}{D} \quad\quad \frac{1}{D^2} \quad\quad e^{-D}$$
 where $D = \lvert \textbf{n}\cdot x - d \rvert$. The exponential kernel $e^{-D}$ corresponds to standard Gaussian assumptions, while $\frac{1}{D}$ and $\frac{1}{D^2}$ represent heavy-tailed distributions. The examples of those methods are demonstrated in Fig.\ref{fig:weight_matrix_computation} and Tab.\ref{tab:weight_matrix}. Empirically, we observed that the inverse-square kernel ($\frac{1}{D^2}$) provides the best balance between convergence speed and robustness of hyperplane fitting. Therefore, we adopted $\frac{1}{D^2}$ as the default kernel for all reported experiments in the main text.

\begin{table}[h]
\label{tab:weight_matrix}
\caption{Result of weight matrix, while our algorithm is set as Sec.5 + Full Pipeline. $\delta = 0.3m$, the total number of points is 1200.}
\begin{center}
\begin{tabular}{lccccccccc}
\toprule
 & \multicolumn{3}{c}{3 hyperplanes} & \multicolumn{3}{c}{4 hyperplanes} & \multicolumn{3}{c}{5 hyperplanes}\\
Kernel & HN & TC & HE & HN & TC & HE & HN & TC & HE\\
\midrule
Gaussian & 3.0 & 305.085 & 1.450 & 4.0 & 314.516 & 2.412 & 5.0 & 331.7 & 3.905\\
$1 / D$ & 3.0 & 794.821 & 6.067 & 4.0 & 697.498 & 7.989 & 5.0 & 668.289 & 9.888\\
$1 / D^2$ & 3.0 & 173.418 & 0.076 & 4.0 & 173.351 & 0.137 & 5.0 & 166.409 & 0.167\\
\bottomrule
\end{tabular}
\end{center}
\end{table}

\subsection{Scaling Experiments}

In this part, we conducted scaling tests of our method. The total number of points in the data instance increases from 600 to 120000, while the number of points generated from the hyperplane is the same. We conduct 10 attempts to test our method, while there are 20 instances in each one. Tab.\ref{tab:2d_scaling} and Tab.\ref{tab:3d_scaling} are results with $\textbf{dim}=2$ and $\textbf{dim}=3$ respectively.

We can find that $\frac{TC}{Total~Number~of~Points} \approx \frac{1}{2}\delta$, i.e., our method can get the global optimal in the scaling test. Moreover, as the number of points increases, the $HE$ decreases since the distribution shifts are alleviated.

  \begin{table*}[h]
\caption{Results of scaling test in 2D data set, while our algorithm is set as Sec.5 + Full Pipeline. $\delta = 0.1m, W = 4\delta$. We can find that $\frac{TC}{Total~Number~of~Points} \approx 0.05$.}
\label{tab:2d_scaling}
\vskip 0.15in
\begin{center}
\begin{small}
\begin{sc}
\begin{tabular}{c|ccc|ccc|ccc}
\toprule
\multirow{2}{*}{Total Number of Points} &
\multicolumn{3}{c|}{3 Hyperplanes} & \multicolumn{3}{c|}{4 Hyperplanes} & \multicolumn{3}{c}{5 Hyperplanes} \\
\cmidrule{2-10}
~ & HN & TC & HE & HN & TC & HE & HN & TC & HE \\
\midrule
 600 &  3.0 & 29.549 & 0.023 & 4.0 & 29.197 & 0.033 & 5.0 & 29.103 & 0.046 \\
 1200 &  3.0 & 59.064 & 0.017 & \textit{3.9} & \textit{81.571} & \textit{0.127} & 5.0 & 58.734 & 0.034 \\
 3000 &  3.0 & 146.470 & 0.008 & 4.0 & 147.781 & 0.014 & 5.0 & 146.236 & 0.019 \\
 9000 & 3.0 & 444.184 & 0.004 & 4.0 & 440.757 & 0.007 & 5.0 & 452.372 & 0.015 \\
 12000 &  3.0 & 564.820 & 0.005 & 4.0 & 588.033 & 0.006 & 5.0 & 589.326 & 0.013 \\
 60000 & 3.0 & 2970.976 & 0.003 & 4.0 & 2946.328 & 0.005 & 5.0 & 2908.838 & 0.012 \\
 120000 & 3.0 & 5899.619 & 0.002 & 4.0 & 5875.969 & 0.003 & 5.0 & 5918.725 & 0.005 \\
\bottomrule
\end{tabular}
\end{sc}
\end{small}
\end{center}
\vskip -0.1in
\end{table*}

\begin{table*}[h]
\caption{Results of scaling test in 3D data set, while our algorithm is set as Sec.5 + Full Pipeline. $\delta = 0.3m, W = 2\delta$. We can find that $\frac{TC}{Total~Number~of~Points} \approx 0.15$.}
\label{tab:3d_scaling}
\vskip 0.15in
\begin{center}
\begin{small}
\begin{sc}
\begin{tabular}{c|ccc|ccc|ccc}
\toprule
\multirow{2}{*}{Total Number of Points} &
\multicolumn{3}{c|}{3 Hyperplanes} & \multicolumn{3}{c|}{4 Hyperplanes} & \multicolumn{3}{c}{5 Hyperplanes} \\
\cmidrule{2-10}
~ & HN & TC & HE & HN & TC & HE & HN & TC & HE \\
\midrule
 600 &  3.0 & 87.403 & 0.091 & 4.0 & 84.397 & 0.138 & 5.0 & 85.92 & 0.165 \\
 1200 &  3.0 & 175.88 & 0.052 & 4.0 & 171.413 & 0.08 & 5.0 & 170.444 & 0.152 \\
 3000 &  3.0 & 439.668 & 0.032 & 4.0 & 434.473 & 0.054 & 5.0 & 427.606 & 0.09 \\
 9000 & 3.0 & 1327.069 & 0.021 & 4.0 & 1303.01 & 0.038 & 5.0 & 1288.649 & 0.055 \\
 12000 &  3.0 & 1754.187 & 0.015 & 4.0 & 1744.676 & 0.031 & 5.0 & 1743.028 & 0.043 \\
 60000 & 3.0 & 8768.782 & 0.013 & 4.0 & 8621.357 & 0.023 & 5.0 & 8597.319 & 0.036 \\
 120000 & 3.0 & 17647.273 & 0.008 & 4.0 & 17826.531 & 0.017 & 5.0 & 17782.893 & 0.029 \\
\bottomrule
\end{tabular}
\end{sc}
\end{small}
\end{center}
\vskip -0.1in
\end{table*}

\subsection{Parameters Robust Experiments}
In this section, we execute the robust test of our method with respect to the parameter $W$. The total number of points in the data instance is 1200, while the number of points generated from the hyperplane is the same. We conduct 10 attempts to test our method, while there are 20 instances in each one. The result is detailed in Tab.\ref{tab:robust_W}.

We can find that our methods are robust with various $W$. While $W$ is near or greater than $2\delta$, our method usually gets the global optimal. However, the number of hyperplanes would be overestimated if $W$ is small.

\begin{table*}[h]
\caption{Results of robust test in 2D data set, while our algorithm is set as Sec.5 + Full Pipeline. $\delta = 0.3m$, the total number of points is 1200.}
\label{tab:robust_W}
\vskip 0.15in
\begin{center}
\begin{small}
\begin{sc}
\begin{tabular}{c|ccc|ccc|ccc}
\toprule
\multirow{2}{*}{Parameter $W$} &
\multicolumn{3}{c|}{3 Hyperplanes} & \multicolumn{3}{c|}{4 Hyperplanes} & \multicolumn{3}{c}{5 Hyperplanes} \\
\cmidrule{2-10}
~ & HN & TC & HE & HN & TC & HE & HN & TC & HE \\
\midrule
 0.1 &  9.4 & 58.898 & 1.384 & 10.0 & 79.667 & 2.108 & 10.0 & 108.622 & 3.445 \\
 0.2 &  5.9 & 90.357 & 0.869 & 8.45 & 82.524 & 1.388 & 9.1 & 97.129 & 1.414 \\
 0.4 &  3.0 & 174.72 & 0.042 & 4.0 & 169.475 & 0.1 & 5.0 & 168.764 & 0.266 \\
 0.6 ($2\delta$) & 3.0 & 174.717 & 0.041 & 4.0 & 169.475 & 0.1 & 5.0 & 168.766 & 0.266 \\
 0.8 &  3.0 & 174.72 & 0.042 & 4.0 & 169.475 & 0.1 & 5.0 & 168.766 & 0.266 \\
 1.0 & 3.0 & 174.72 & 0.042 & 4.0 & 169.475 & 0.1 & 5.0 & 168.766 & 0.266 \\
 1.2 ($4\delta$) & 3.0 & 174.72 & 0.042 & 4.0 & 169.475 & 0.1 & 4.9 & 171.474 & 0.577 \\
\bottomrule
\end{tabular}
\end{sc}
\end{small}
\end{center}
\vskip -0.1in
\end{table*}

\subsection{Test with Solver} \label{compare_solver}

To simplify the distance computation between points hyperplane, we use different features to represent the hyperplane and test it with Gurobi\cite{gurobi}. The result is shown in Tab.\ref{tab:distance}.

\begin{table}[h]
\caption{Results of Gurobi. There is one line in 2D space. The black points are data, the dotted gray lines are ground truth, and the dotted blue lines are the result of the solver.}
\label{tab:distance}
\vskip 0.15in
\begin{center}
\begin{small}
\begin{sc}
\begin{tabular}{lcc}
\toprule
Representation & Result & Comment \\
\midrule
Normal Vector and Distance \\ $\bm{\textup{n}} \cdot x - d = 0, \lVert \bm{\textup{n}} \rVert_2 = 1$  & \begin{minipage}[b]{0.15\columnwidth}
\centering
\raisebox{-.5\height}{\includegraphics[width=\linewidth]{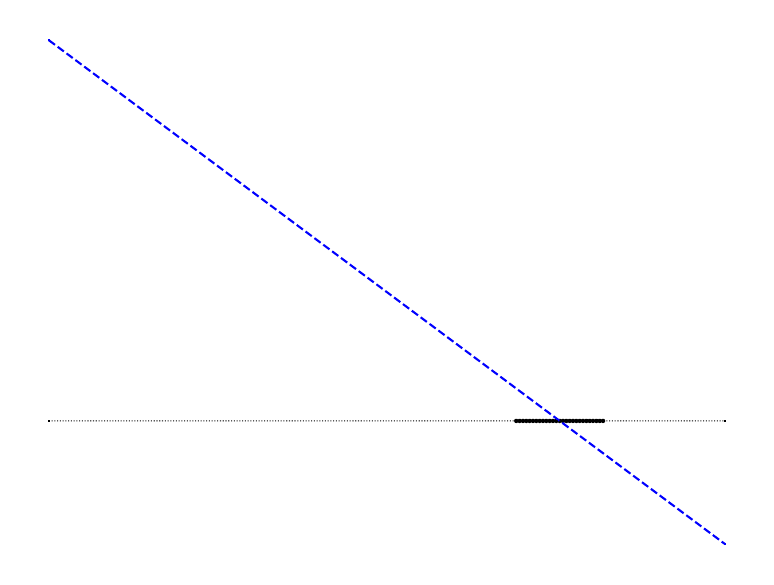}}
\end{minipage} & \textup{not work well with non-convex constraint}\\
\midrule
Normal Vector and Point on Hyperplane \\ $\bm{\textup{n}} \cdot (x - p) = 0, \lVert \bm{\textup{n}} \rVert_2 = 1$  & \begin{minipage}[b]{0.15\columnwidth}
\centering
\raisebox{-.5\height}{\includegraphics[width=\linewidth]{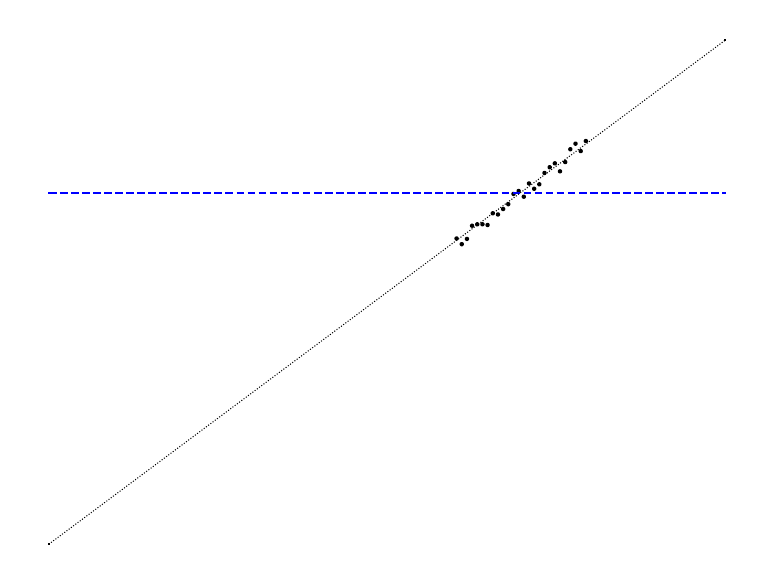}}
\end{minipage} & \textup{not work well with non-convex constraint} \\
\midrule
Theta and Point on Hyperplane \\ $\bm{\textup{n}} \cdot (x - p) = 0, \bm{\textup{n}} = [\cos\theta, \sin\theta]$    & \begin{minipage}[b]{0.15\columnwidth}
\centering
\raisebox{-.5\height}{\includegraphics[width=\linewidth]{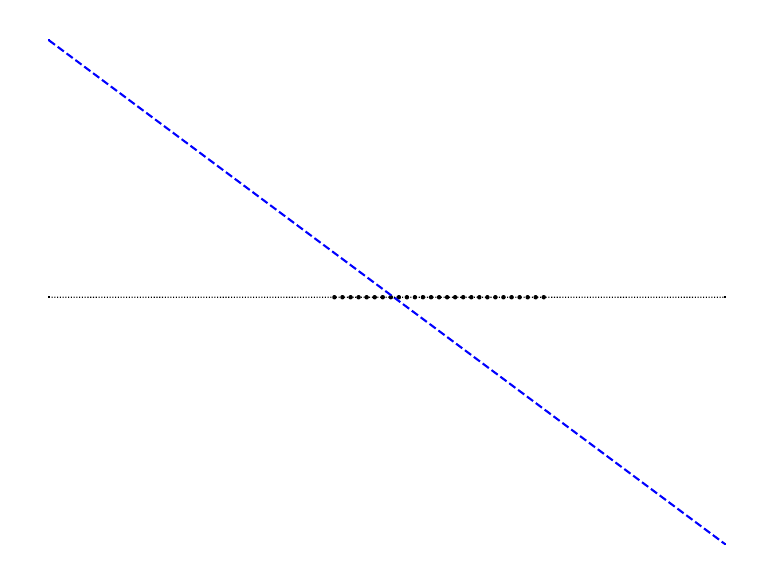}}
\end{minipage} & \textup{Computationally expensive or infeasible}\\
\midrule
Only Normal Vector, Minimum Variance \\ $\min \frac{1}{n} [\sum(\bm{\textup{n}}\cdot x)^2 - \frac{1}{n} (\sum\bm{\textup{n}}\cdot x)^2]$     & \begin{minipage}[b]{0.15\columnwidth}
\centering
\raisebox{-.5\height}{\includegraphics[width=\linewidth]{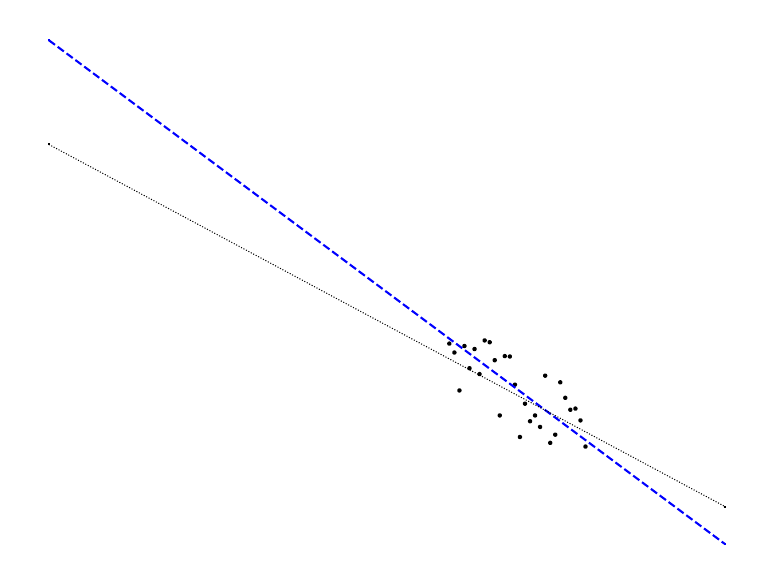}}
\end{minipage} & \textup{get local optimum, maybe infeasible}\\

\bottomrule
\end{tabular}
\end{sc}
\end{small}
\end{center}
\vskip -0.1in
\end{table}

In summary, the commercial solver can not perform well on a non-convex problem, easily getting a local optimum. And it costs a lot or results in an infeasible solution where there are non-convex constraints.

\clearpage

\section{Appendix Part III}
\subsection{Results Visualization} \label{result_visualization}
\begin{figure}[h]
\vskip 0.2in
\begin{center}
\includegraphics[width=0.9\textwidth]{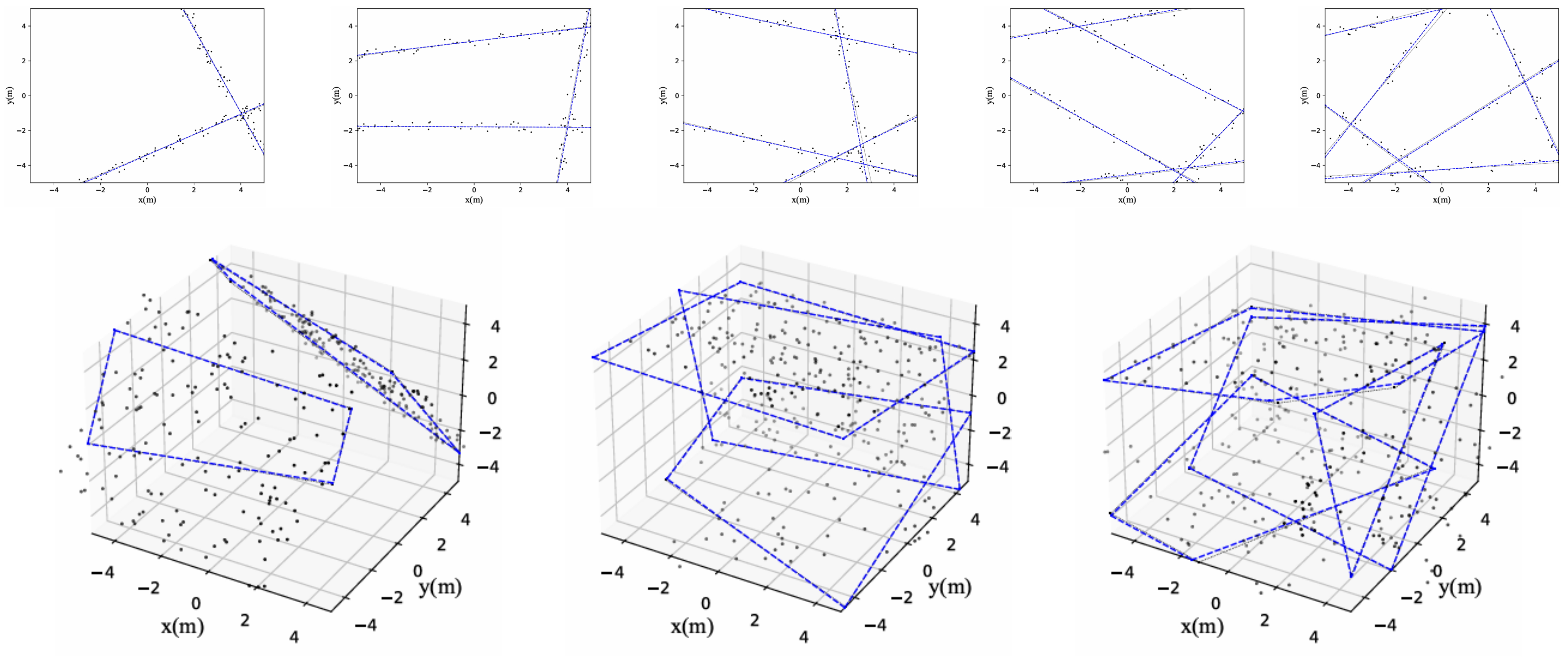}
\caption{Some results of our algorithms in 2D and 3D space, $\delta = 0.3m$. Our algorithm can find all hyperplanes.}
\label{fig:result}
\end{center}
\vskip -0.2in
\end{figure}

Here we present some visual results of our method in various data instances.
Fig.\ref{fig:result} shows some results in 2D and 3D space. 

\subsection{Potential Application}
When deploying our method in practical engineering applications, several adaptations can be made accordingly:
\begin{itemize}
\item For hyperplane detection in point clouds, if it is not required to assign all points to the hyperplane, a distance threshold can be applied prior to Hard Refinement to eliminate points whose distance to the nearest hyperplane exceeds this threshold. In this way, outliers will not be involved in estimating the final hyperplane during the Hard Refinement stage.
\item If the task only requires hyperplane-based point classification instead of estimating the ground-truth hyperplane, the Hard Refinement module can be removed from the pipeline. After Soft Optimization converges, points with distances to the nearest hyperplane exceeding the threshold are discarded, and the class assignment label for each point can be directly obtained.
\item For applications with coarser precision requirements, the hyperplane derived from the Initial Value stage alone is sufficient to perform point classification.
\end{itemize}

\end{document}